\newcommand{\tabincell}[2]{\begin{tabular}{@{}#1@{}}#2\end{tabular}}
\newcommand{\revision}[1]{{#1}}

\newcommand{\eg}{e.g.}
\newcommand{\ie}{i.e.}
\newcommand{\etc}{etc}
\newcommand{\etal}{et al.}

\documentclass[preprint,review,12pt]{elsarticle}

\usepackage{amssymb, amsmath}
\usepackage{subfigure, booktabs}
\usepackage{color}
\journal{Pattern Recognition}

\begin{document}

\begin{frontmatter}



\title{A Comprehensive Evaluation Framework for Deep Model Robustness}


\author[label1]{Jun Guo}
\author[label2]{Wei Bao}
\author[label5]{Jiakai Wang}
\author[label1]{Yuqing Ma}
\author[label3]{Xinghai Gao}
\author[label4]{Gang Xiao}
\author[label1]{Aishan Liu\corref{1}}
\ead{liuaishan@buaa.edu.cn}
\author[label1,label2]{Jian Dong}
\author[label1]{Xianglong Liu\corref{1}}
\ead{xlliu@buaa.edu.cn}
\author[label1]{Wenjun Wu}

\affiliation[label1]{organization={State Key Lab of Software Development Environment},
            addressline={Beihang University}, 
            city={Beijing},
            country={China}}
        
\affiliation[label2]{organization={China Electronics Standardization Institute},
            city={Beijing},
            country={China}}
            
\affiliation[label5]{organization={Zhongguancun Laboratory},
            city={Beijing},
            country={China}}
            
\affiliation[label3]{organization={Institute of Unmanned System},
            addressline={Beihang University}, 
            city={Beijing},
            country={China}}
            
\affiliation[label4]{organization={National Key Laboratory for Complex Systems Simulation},
            city={Beijing},
            country={China}}

\cortext[1]{Corresponding author.}
            
\begin{abstract}
Deep neural networks (DNNs) have achieved remarkable performance across a wide range of applications, while they are vulnerable to adversarial examples, which motivates the evaluation and benchmark of model robustness. However, current evaluations usually use simple metrics to study the performance of defenses, which are far from understanding the limitation and weaknesses of these defense methods. Thus, most proposed defenses are quickly shown to be attacked successfully, which results in the ``arm race'' phenomenon between attack and defense. To mitigate this problem, we establish a model robustness evaluation framework containing 23 comprehensive and rigorous metrics, which consider two key perspectives of adversarial learning (i.e., data and model). Through neuron coverage and data imperceptibility, we use data-oriented metrics to measure the integrity of test examples; by delving into model structure and behavior, we exploit model-oriented metrics to further evaluate robustness in the adversarial setting. To fully demonstrate the effectiveness of our framework, we conduct large-scale experiments on multiple datasets including CIFAR-10, SVHN, and ImageNet using different models and defenses with our open-source platform. Overall, our paper provides a comprehensive evaluation framework, where researchers could conduct comprehensive and fast evaluations using the open-source toolkit, and the analytical results could inspire deeper understanding and further improvement to the model robustness.
\end{abstract}


\begin{keyword}
 Adversarial examples \sep Evaluation metrics \sep Model robustness
\end{keyword}

\end{frontmatter}


% Main text
\section{Introduction}\label{Section:introduction}

Deep learning models have achieved remarkable performance across a wide range of applications, however, they are susceptible to \emph{adversarial examples} {\cite{szegedy2013intriguing}}. Since deep learning has been integrated into various {safety-critical scenarios}, the safety problem brought by adversarial examples has attracted extensive attention from the perspectives of both {adversarial attack\cite{goodfellow6572explaining,athalye2017synthesizing} and defense \cite{dai2022deep,madry2017towards,liu2019training}}. 
Evaluating and benchmarking the robustness of deep learning models, as a direct and effective approach, paves a very fundamental path to better understanding and further improving model robustness {\cite{carlini2019evaluating,Ling2019Deepsec,Ma2018DeepGauge}}.
However, most of these works focus on providing practical advice or benchmarking the performance of adversarial defenses, which ignore the significance of evaluation metrics. By adopting the simple evaluation metrics (e.g., attack success rate, classification accuracy), most of the current studies could only use model outputs to conduct incomplete evaluations, which fail to provide comprehensive understandings of the limitations of these defenses. Thus, these defenses are quickly shown to be attacked successfully, which results in the ``arm race'' phenomenon between attacks and defenses. Therefore, it is of great significance and challenge to conduct rigorous and extensive evaluation on robustness for navigating the research field and further facilitating trustworthy deep learning in practice.

\begin{figure*}[!t]
	\centering
	\includegraphics[width=0.8\linewidth]{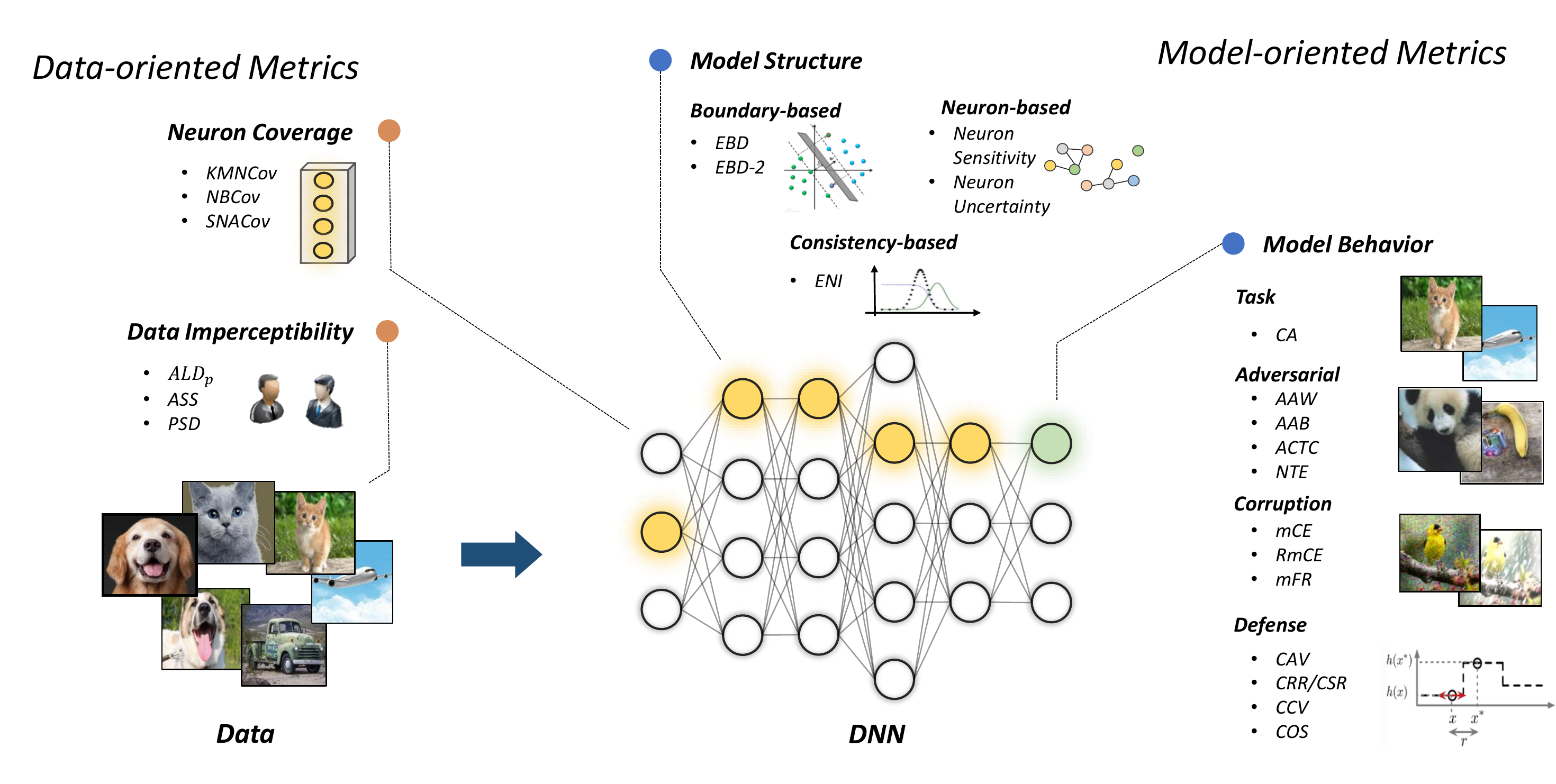}
	\caption{With 23 evaluation metrics in total, our comprehensive evaluation framework primarily focuses on the two key factors of adversarial learning (i.e., data and model).}
	\label{fig:framework}
\end{figure*}

In this work, with a hope to facilitate future research, we establish a model robustness evaluation framework containing a comprehensive, rigorous, and coherent set of evaluation metrics. These metrics could fully evaluate model robustness and provide deep insights into building robust models. This paper focuses on the robustness of deep learning models on the most commonly studied image classification tasks with respect to $\ell_p$-norm bounded adversaries and some other corruption. As illustrated in Figure \ref{fig:framework}, our evaluation framework can be roughly divided into two parts: data-oriented and model-oriented, which focus on the two key factors of adversarial learning  (i.e., data and model). {Data-oriented metrics focus on the integrity of test examples (i.e., whether the conducted evaluation covers most of the neurons within a model), while model-oriented metrics consider both model structures (\eg, neurons, layers) and behaviors in the adversarial setting (\eg, adversarial performance, decision boundary).} Our framework contains 23 evaluation metrics in total. 

% Since model robustness is evaluated based on a set of perturbed examples, we first use data-oriented metrics regarding neuron coverage and data imperceptibility to measure the integrity of test examples (i.e., whether the conducted evaluation covers most of the neurons within a model); meanwhile, we focus on evaluating model robustness via model-oriented metrics which consider both model structures and behaviors in the adversarial setting (e.g., decision boundary, model neuron, corruption performance, etc.). 

To fully demonstrate the effectiveness of the evaluation framework, we then conduct large-scale experiments on multiple datasets (i.e., CIFAR-10 \cite{krizhevsky2009learning}, SVHN \cite{netzer2011reading}, and ImageNet \cite{deng2009imagenet}) using different models with different adversarial defense strategies. Through the experimental results, we could conclude that: (1) though showing high performance on some simple and intuitive metrics such as adversarial accuracy, some defenses are weak on more rigorous and insightful metrics; (2) besides $\ell_{\infty}$-norm adversarial examples, more diversified attacks should be performed to conduct comprehensive evaluations (e.g., corruption attacks, $\ell_2$ adversarial attacks, etc.); (3) apart from model robustness evaluation, the proposed metrics shed light on the model robustness and are also beneficial to the design of adversarial attacks and defenses. All evaluation experiments are conducted on our new adversarial robustness evaluation platform, which we hope could facilitate follow researchers for a better understanding of adversarial examples as well as further improvement of model robustness. Our contributions are as follows:
\begin{itemize}
\item{We establish a comprehensive evaluation framework for model robustness containing 23 data-oriented and model-oriented metrics, which could fully evaluate model robustness through static structure and dynamic behavior, and provide deep insights into building robust models;}

\item{Based on our framework, we provide an open-sourced platform, which supports continuous integration of user-specific algorithms and language-independent models;}

\item{We conduct large-scale experiments, and we provide preliminary suggestions to the design of adversarial attacks/defenses in the future. {Meanwhile, we provide suggestions on the selection of proper metrics with examples.}}

\end{itemize}

\section{Related Work} 
\label{Section:relatedwork}

\subsection{Adversarial attacks and defenses}
Adversarial examples are inputs intentionally designed to mislead DNNs \cite{szegedy2013intriguing,goodfellow6572explaining}. Given a DNN $f$ and an input image $\mathbf{x} \in \mathbb X$ with the ground truth label $\mathbf y \in \mathbb Y$, an adversarial example $\mathbf{x}_{adv}$ satisfies
\begin{align*}
f(\mathbf x_{adv}) \neq \mathbf y  \quad s.t. \quad \|\mathbf x-\mathbf x_{adv}\| \leq \epsilon,
\end{align*}
where $\|\cdot\|$ is a distance metric measured by the $\ell_{p}$-norm ($p\in$\{1,2,$\infty$\}).

In the past years, great efforts have been devoted to generating adversarial examples in different scenarios and tasks {\cite{Liu2020Spatiotemporal,Liu2020Biasbased}}.
Adversarial attacks can be divided into two types: white-box attacks, in which adversaries have the complete knowledge of the target model and can fully access the model {\cite{goodfellow6572explaining,madry2017towards,Carlini2017towards,shi2020adaptive}}; black-box attacks, in which adversaries have limited knowledge of the target classifier and can not directly access the model {\cite{hang2020ensemble,Wang2022ViTAttack,liang2021parallel}}.

Meanwhile, to improve model robustness against adversarial examples, various defense approaches have been proposed, including defensive distillation \cite{Papernot2015Distillation}, input transformation {\cite{xie2018mitigating}}, robust training \cite{madry2017towards}, and certified defense \cite{Croce2020Provable}, among which adversarial training has been widely studied and demonstrated to be the most effective \cite{goodfellow6572explaining,madry2017towards}. Besides, corruption such as snow and blur also frequently occur in the real world, which also presents critical challenges for the building of robust deep learning models. Average-case model performance on small, general, classifier-agnostic corruption can be used to define model corruption robustness.

\subsection{Model robustness evaluation}

{Most proposed defenses conduct incomplete or incorrect evaluations, which are quickly shown to be attacked successfully due to limited understanding of these defenses {\cite{Carlini2016Defensive,Carlini2017towards,athalye2018obfuscated}}. Consequently, conducting rigorous and comprehensive evaluation on model robustness becomes particularly important.} To comprehensively evaluate the model robustness for DNNs, a number of works have been proposed. A uniform platform for adversarial robustness analysis named DEEPSEC \cite{Ling2019Deepsec} is proposed to measure the vulnerability of deep learning models. Specifically, the platform incorporates 16 adversarial attacks with 10 attack utility metrics and 13 adversarial defenses with 5 defensive utility metrics. Unlike prior works, \cite{carlini2019evaluating} discussed the methodological foundations, reviewed commonly accepted best practices, and suggested new methods for evaluating defenses to adversarial examples. In particular, they provided principles for performing defense evaluations and a specific checklist for avoiding common evaluation pitfalls. Moreover, \cite{Ma2018DeepGauge} proposed a set of multi-granularity metrics for deep learning systems, which aims at rendering a multi-faceted portrayal of the testbed (i.e., testing coverage). More recently, {RobustART\cite{tang2021robustart} and RobustBench\cite{croce2020robustbench} have proposed large-scale benchmarks for evaluating adversarial robustness in terms of adversarial attacks, defenses and model structures, but they only utilize accuracy as the evaluation metric.} 

However, these studies mainly focus on establishing open-source libraries for attacks/defenses, which fail to provide a comprehensive evaluation considering several aspects of a deep learning model towards different noises.

\begin{table}[!htb]
\centering
\caption{The taxonomy and illustration of the proposed evaluation metrics.}
\vspace{-0.2in}
\begin{center}
\begin{scriptsize}
\begin{sc}
%\scriptsize
%\setlength{\tabcolsep}{0.2mm}{
%\begin{tabular}{cc|cccccccc}
\resizebox{\linewidth}{!}{
    \begin{tabular}{c|cccccccc}
    \toprule
    Metrics & Behavior & structure & \tabincell{c}{Adversarial\\attacks} & \tabincell{c}{Corruption\\attacks} &  Whitebox & Blackbox & \tabincell{c}{Single\\model} & \tabincell{c}{Multiple\\models}\\
    \hline
    %\multirow{6}{*}{Data}
    KMNCov \cite{Ma2018DeepGauge} &  &   & \checkmark & \checkmark & \checkmark & &\checkmark & \\
    NBCov \cite{Ma2018DeepGauge}&  &   &  \checkmark & \checkmark & \checkmark & &\checkmark &\\
    SNACov \cite{Ma2018DeepGauge}&  &   & \checkmark & \checkmark & \checkmark & &\checkmark &\\
    ALD$_p$ \cite{Ling2019Deepsec}&  &   & \checkmark & & & \checkmark &\checkmark &\\
    ASS \cite{ass}&  &   & \checkmark & & & \checkmark &\checkmark &\\
    PSD \cite{Luo_Liu_Wei_Xu_2018}&  &   & \checkmark & & & \checkmark &\checkmark &\\
    \hline
    %\multirow{17}{*}{Model} &
    CA & \checkmark &  & & & & \checkmark &\checkmark & \\
    AAW & \checkmark &  & \checkmark & & \checkmark & &\checkmark & \\
    AAB & \checkmark &  & \checkmark & & & \checkmark & & \checkmark\\
    ACAC \cite{acac}& \checkmark &  & \checkmark & & \checkmark & & \checkmark &\\
    ACTC \cite{acac}& \checkmark &  & \checkmark & & \checkmark & & \checkmark &\\
    NTE \cite{Luo_Liu_Wei_Xu_2018}& \checkmark &  & \checkmark& & \checkmark & & \checkmark &\\
    mCE \cite{hendrycks2018benchmarking}& \checkmark &  & & \checkmark & & \checkmark & \checkmark &\\
    RmCE \cite{hendrycks2018benchmarking}& \checkmark &  & & \checkmark & & \checkmark & & \checkmark\\
    mFR \cite{hendrycks2018benchmarking}& \checkmark &  & & \checkmark & & \checkmark & & \checkmark\\
    CAV \cite{Ling2019Deepsec}& \checkmark &  & \checkmark & \checkmark & & \checkmark & & \checkmark\\
    CRR/CSR \cite{Ling2019Deepsec}& \checkmark &  & \checkmark & \checkmark & & \checkmark & & \checkmark\\
    CCV \cite{Ling2019Deepsec}& \checkmark &  & \checkmark & \checkmark & \checkmark & & & \checkmark\\
    COS \cite{Ling2019Deepsec}& \checkmark &  & \checkmark & \checkmark & \checkmark & & & \checkmark\\
    EBD \cite{liu2019training}&  & \checkmark & \checkmark & \checkmark & \checkmark & & \checkmark&\\
    EBD-2 &  & \checkmark & \checkmark & & \checkmark & & \checkmark& \\
    ENI  \cite{liu2019training}&  & \checkmark & \checkmark& \checkmark&  & \checkmark & \checkmark&\\
    Neuron Sensitivity \cite{9286885} &  & \checkmark &  \checkmark & & \checkmark & & \checkmark&\\
    Neuron Uncertainty &  & \checkmark &\checkmark &\checkmark &  \checkmark & & \checkmark& \\
    %\hline
    \bottomrule
    \end{tabular}
}
\end{sc}
\end{scriptsize}
\end{center}
\label{table:allmetrics}
\end{table}

\section{Evaluation Metrics}
\label{Section:metrics}
To mitigate the problem brought by incomplete evaluation, we establish a multi-view model robustness evaluation framework which consists of 23 evaluation metrics in total. As shown in Table \ref{table:allmetrics}, our evaluation metrics can be roughly divided into two parts: data-oriented and model-oriented.

\subsection{Data-Oriented Evaluation Metrics}
Since model robustness is evaluated based on a set of perturbed examples, the quality of the test data plays a critical role in robustness evaluation. Thus, we use data-oriented metrics considering both neuron coverage and data imperceptibility to measure the integrity of test examples. DeepGauge \cite{Ma2018DeepGauge} introduced coverage criteria into neural networks and proposed Neuron Coverage to leverage the output values of neuron and its corresponding boundaries obtained from training data to approximate the major function region and the corner-case region at the neuron level.

\subsubsection{\textbf{Neuron Coverage}} We first use the coverage criteria for DNNs to measure whether the generated test set could cover enough amount of neurons.

\emph{$k$-Multisection Neuron Coverage (KMNCov).} Given a neuron $\mathbf{n}$, the KMNCov measures how thoroughly the given set of the test inputs $\mathbb{D}$ covers the range of neuron output value [${low}_\mathbf{n}$, ${high}_\mathbf{n}$], where $\mathbb{D}=\{x_1, x_2,...\}$ is a set of input data. Specifically, we divide the range [${low}_\mathbf{n}$, ${high}_\mathbf{n}$] into $k$ sections with the same size ($k > 0$), and $S_i^\mathbf{n}$ denotes the $i$-th section where $1 \leq i \leq k$. Let $\phi(x,\mathbf{n})$ denote a function that returns the output of a neuron $\mathbf{n}$ under a given input sample $x \in \mathbb{D}$. We use $\phi(x,\mathbf{n}) \in S_i^\mathbf{n}$ to denote that the $i$-th section of neuron $n$ is covered by the input $x$. For a given test set $\mathbb{D}$ and a specific neuron $\mathbf{n}$, the corresponding $k$-Multisection Neuron Coverage is defined as the ratio of the sections covered by $\mathbb{D}$ and the overall sections as
\begin{equation}
    {\rm {KMNCov}}(\mathbb{D},k) = \frac{\sum_{\mathbf{n}\in N}|\{S_i^\mathbf{n}|\exists  x \in \mathbb{D} : \phi(x, \mathbf{n})\in S_i^\mathbf{n}\}}{k\times |N|},
\end{equation}
where $N = \{\mathbf{n}_1,\mathbf{n}_2,...\}$ is a set of neurons for the model. It should be noticed that for a neuron $n$ and input $x$, if $\phi (x,\mathbf{n}) \in [low_\mathbf{n}, high_\mathbf{n}]$ is satisfied with $\forall \mathbf{n}\in N$, we say that this DNN is located in its major function region. Otherwise, it is located in the corner-case region. \revision{In our experiments, we set the number of sections $k$ as 100, and [${low}_\mathbf{n}$, ${high}_\mathbf{n}$] is determined by the activation value of neuron $\mathbf{n}$ on the training set.} {KMNCov reflects the comprehensiveness of the test set. A qualified test set should have a high value of KMNCov, which means the test set have thoroughly tested the neural network.}

\emph{Neuron Boundary Coverage (NBCov)}. It measures how many corner-case regions have been covered by the given test input set $\mathbb{D}$. Given an input $\mathbf{x}\in \mathbb{D}$, a DNN is located in its corner-case region when given $\mathbf{x}$, $ \exists \mathbf{n} \in N :$ $\phi(\mathbf{x},\mathbf{n}) \in (-\infty, {low}_{\mathbf{n}})\cup ({high}_{\mathbf{n}}, \infty)$. Thus, the NBCov can be defined as the ratio of the covered corner cases and the total corner cased ($2\times \|N\|$):

\begin{equation}
\begin{aligned}
    \frac{|UpperCornerNeuron| + |LowerCornerNeuron|}{2\times|N|},
\end{aligned}
\end{equation}
where the $UpperCornerNeuron$ is the set consisting of neurons that satisfy $\exists \mathbf{x}\in \mathbb{D}: \phi(\mathbf{x},n)\in({high}_\mathbf{n},+\infty)$. And the $LowerCornerNeuron$ is the set of neurons that satisfy $\exists \mathbf{x}\in \mathbb{D}:\phi(\mathbf{x},\mathbf{n})\in (-\infty, {low}_\mathbf{n})$.

\emph{Strong Neuron Activation Coverage (SNACov)}. This metric is designed to measure the coverage status of upper-corner case (i.e., how many corner cases have been covered by the given test sets). It can be described as the ratio of the covered upper-corner cases and the total corner cases ($|N|$):
\begin{equation}
  \begin{aligned}
    {SNACov}(\mathbb{D}) = \frac{|UpperCornerNeuron|}{|N|}.
  \end{aligned}
\end{equation}
{According to the coverage criteria, model robustness is closely related to the number of corner cases. High NBCov and SNACov indicate the model meets many unexpected inputs and cannot handle them well.}

\subsubsection{\textbf{Data Imperceptibility}}
Here, we introduce several metrics to evaluate data visual imperceptibility by considering the magnitude of perturbations.

\emph{Average $\ell_p$ Distortion (ALD$_p$).} Most adversarial attacks generate adversarial examples by constructing additive $\ell_p$-norm adversarial perturbations (e.g., $p \in {0,1,...,\infty}$). To measure the visual perceptibility of generated adversarial examples, we use ALD$_p$ as the average normalized $\ell_p$ distortion:
\begin{equation}
\begin{aligned}
ALD_p=\frac{1}{m}\sum_{i=1}^{m}{\frac{||\mathbf{x}^{(i)}_{adv}-\mathbf{x}^{(i)}||_p}{||\mathbf{x}^{(i)}||_p}},
\end{aligned}
\end{equation}
where $m$ denotes the number of adversarial examples that attack successfully, and the smaller ALD$_p$ is, the more imperceptible the adversarial example is.

\emph{Average Structural Similarity (ASS).} To evaluate the imperceptibility of adversarial examples, we further use SSIM which is considered to be effective to measure human visual perception. {SSIM \cite{SSIM2004Zhou} is the most commonly used metric to evaluate the structure similarity between two images. It separates the task of similarity measurement into three comparisons: luminance, contrast, and structure as} 
\begin{equation}
\begin{aligned}
SSIM(x,y)=\frac{(2\mu_x\mu_y+c_1)(2\sigma_{xy}+c_2)}{(\mu^2_x+\mu^2_y+c_1)(\sigma^2_x+\sigma^2_y+c_2)}
\end{aligned}
\end{equation}
{where $\mu_x$ and $\mu_y$ represent the means of input image $x$ and $y$, $\sigma^2_x$ and $\sigma^2_y$ stand for the variances of $x$ and $y$, $\sigma_{xy}$ is the covariance of $x$ and $y$, $c_1$ and $c_2$ are constants. In our experiments, we set $c_1=0.01$ and $c_2=0.03$. Thus, ASS can be defined as the average of SSIM between all adversarial examples and the corresponding clean examples, i.e.,}
\begin{equation}
\begin{aligned}
ASS=\frac{1}{m}\sum_{i=1}^{m}{SSIM(\mathbf{x}^{(i)}_{adv},\mathbf{x}^{(i)})},
\end{aligned}
\end{equation}
where $m$ denotes the number of successful adversarial examples, and the higher ASS is, the more imperceptible the adversarial example is.

\emph{Perturbation Sensitivity Distance (PSD).} Based on the contrast masking theory \cite{legge1980contrast}, PSD is proposed to evaluate human perception of perturbations. Thus, PSD is defined as:
\begin{equation}
\begin{aligned}
PSD=\frac{1}{m}\sum_{i=1}^{m}\sum_{i=1}^{t}{\delta^{(i)}_{(j)}Sen(R(\mathbf{x}^{(i)}_{(j)}))},
\end{aligned}
\end{equation}
{where $m$ denotes the number of adversarial examples that attacks successfully, $t$ is the total number of pixels,  $x^{(i)}_{(j)}$ represents the $j$-th pixel of the $i$-th example, and $\delta^{(i)}_{(j)}$ represents the perturbations added at the specific pixel}. $ R(\mathbf{x}^{(i)}_{(j)})$ stands for the square surrounding region of $\mathbf{x}^{(i)}_{(j)}$, and $Sen(R(\mathbf{x}^{(i)}_{(j)}))=1/std(R(\mathbf{x}^{(i)}_{(j)}))$. According to contrast masking theory in image processing \cite{legge1980contrast}, human eyes are more sensitive to perturbations on pixels in low variance regions than those in high variance regions. Therefore, to make adversarial examples imperceptible, we should perturb pixels at high variance zones rather than low variance ones. Evidently, the smaller PSD is, the more imperceptible the adversarial example is.

\subsection{Model-oriented Evaluation Metrics}
To evaluate robustness, the most intuitive direction is to measure the model performance in the adversarial setting. Given an adversary $\mathcal{A}_{\epsilon,p}$, it uses specific attack methods to generates adversarial examples $\mathbf{x}_{adv}$ = $\mathcal{A}_{\epsilon,p}(\mathbf{x})$ for a clean example $\mathbf{x}$ with the perturbation magnitude $\epsilon$ under $\ell_p$ norm. 
\subsubsection{\textbf{Model Behaviors}}

\begin{itemize}
\item \textbf{Task Performance}
\end{itemize}

\emph{Clean Accuracy (CA).} Model accuracy on clean examples is one of the most important properties in the adversarial setting. CA is defined as the percentage of clean examples that are successfully classified by a classifier $f$ into the ground truth classes as follows
\begin{equation}
\begin{aligned}
CA(f,\mathbb{D})=\frac{1}{n}\sum_{i=1}^{n}{\mathbf{1}(f(\mathbf{x}_i)=\mathbf{y}_i)},
\end{aligned}
\end{equation}
where $\mathbb{D}=\{\mathbf x^{(i)}, \mathbf y^{(i)}\}_{i=1...n}$ is the test set, $\mathbf{1}(\cdot)$ is the indicator function.

\begin{itemize}
\item \textbf{Adversarial Performance}
\end{itemize}

\emph{Adversarial Accuracy on White-box Attacks (AAW).} In the untargeted attack scenario, AAW is defined as the percentage of adversarial examples generated in the white-box setting that are successfully misclassified into an arbitrary class except for the ground truth class; for targeted attack, it can be measured by the percentage of adversarial examples generated in the white-box setting classified as the target class. In the rest of the paper, we mainly focus on untargeted attacks. Thus, AAW can be defined as:
\begin{equation}
\begin{aligned}
AAW(f,\mathbb{D},\mathcal{A}_{\epsilon,p})=\frac{1}{n}\sum_{i=1}^{n}{\mathbf{1}(f(\mathcal{A}_{\epsilon,p}(\mathbf{x}_i))\neq\mathbf{y}_i)}.
\end{aligned}
\end{equation}

\emph{Adversarial Accuracy on Black-box Attacks (AAB).} Similar to AAW, AAB is defined by the percentage of black-box adversarial examples classified correctly by the classifier.

\emph{Average Confidence of Adversarial Class (ACAC).} Besides prediction accuracy, prediction confidence on adversarial examples gives further indications of model robustness. Thus, for an adversarial example, ACAC can be defined as the average prediction confidence towards the incorrect class
\begin{equation}
\begin{aligned}
ACAC(f,\mathbb{D},\mathcal{A}_{\epsilon,p})=\frac{1}{m}\sum_{i=1}^{m}{P(\mathcal{A}_{\epsilon,p}(\mathbf{x}_i))},
\end{aligned}
\end{equation}
{where $m$ is the number of adversarial examples that attack successfully, $P$ is the prediction confidence of classifier $f$ towards the wrong classes.}

\emph{Average Confidence of True Class (ACTC).} Meanwhile, we also use ACTC to further evaluate to what extent the attacks escape from the ground truth. In other words, ACTC can be defined as the average model prediction confidence on adversarial examples towards the ground truth labels, i.e.,
\begin{equation}
\begin{aligned}
ACTC(f,\mathbb{D},\mathcal{A}_{\epsilon,p})=\frac{1}{m}\sum_{i=1}^{m}{P_{\mathbf{y}_i}(\mathcal{A}_{\epsilon,p}(\mathbf{x}_i))},
\end{aligned}
\end{equation}
{where $m$ is the number of adversarial examples that attack successfully, $P_{\mathbf{y}_i}$ is the prediction confidence of classifier $f$ towards the ground truth class $y_i$.}

\emph{Noise Tolerance Estimation (NTE).} Moreover, given the generated adversarial examples, we further calculate the gap between the probability of misclassified class and the max probability of all other classes, {which measures the tolerance of model against adversarial noises.}
\begin{equation}
\begin{aligned}
NTE(f,\mathbb{D},\mathcal{A}_{\epsilon,p})=\frac{1}{m}\sum_{i=1}^{m}[P_{f(\mathbf{x}_i)}(\mathcal{A}_{\epsilon,p}(\mathbf{x}_i))- \max{P_{j}(\mathcal{A}_{\epsilon,p}(\mathbf{x}_i))}],
\end{aligned}
\end{equation}
where $j \in {1,...,k}$ and $j \neq f(\mathbf{x}_i)$, { and $m$ is the number of adversarial examples that attack successfully. As the NTE becomes increasingly high, the model becomes more vulnerable towards adversarial examples.}

\begin{itemize}
\item \textbf{Corruption Performance}
\end{itemize}

To further comprehensively measure the model robustness against different corruption, we introduce evaluation metrics following \cite{hendrycks2018benchmarking}.

\emph{mCE.} This metric denotes the mean corruption error of a model compared to the baseline model \cite{hendrycks2018benchmarking}. Different from the original paper, we simply calculate the error rate of the classifier $f$ on each corruption type $c$ at each level of severity $s$ denoted as $E^f_{s,c}$ and compute mCE as follows:
\begin{equation}
\begin{aligned}
mCE_c^f = \frac{1}{t} \sum_{s=1}^t E^f_{s,c},
\end{aligned}
\end{equation}
where $t$ denotes the number of severity levels. Thus, mCE is the average value of Corruption Errors (CE) using different corruption.

\emph{Relative mCE.} A more nuanced corruption robustness measure is Relative mCE (RmCE) \cite{hendrycks2018benchmarking}. If a classifier withstands most corruption, the gap between mCE and the clean data error is minuscule. So, RmCE is
\begin{equation}
\begin{aligned}
RmCE_c^f = \frac{1}{t} \sum_{s=1}^t E^f_{s,c} - E^f_{clean},
\end{aligned}
\end{equation}
where $E^f_{clean}$ is the error rate of $f$ on clean examples.

\emph{mFR.} Hendrycks \etal \cite{hendrycks2018benchmarking} introduce mFR to represent the classification differences between two adjacent frames in the noise sequence for a specific image. Let us denote $q$ noise sequences with $S=\{(\mathbf{x}_i^{(1)},\mathbf{x}_i^{(2)},...,\mathbf{x}_i^{(n)})\}_{i=1}^q$ where each sequence is created with a specific noise type $c$ as
\begin{equation}
\begin{aligned}
FP_c^f &= \frac{1}{q(n-1)} \sum_{i=1}^qsum_{j=2}^n \mathbf{1} (f(\mathbf{x}_i^{(j)}) \neq f(\mathbf{x}_{i}^{(j-1)})).
\end{aligned}
\end{equation}
Then, the Flip Rate (FR) can be obtained by $FR_c^f=FP_c^f/FP_c^{base}$ and mFR is the average value of FR. {Different from the original paper, we set $FP_c^{base}$ to be 1, which means $FR_c^f=FP_c^f$.}
%{where $FP_c^{base}$ represents the Flip Probability of a base model (i.e., a vanilla model) towards the noise type $c$.}

\begin{itemize}
\item \textbf{Defense Performance}
\end{itemize}
We further explore to what extent the model performance has been influenced when defense strategies are added.

\emph{CAV.} Classification Accuracy Variance (CAV) is used to evaluate the impact of defenses based on the accuracy. We expect the defense-enhanced model $f^d$ to maintain the classification accuracy on normal testing examples as much as possible. Therefore, it is defined as follows:
\begin{equation}
\begin{aligned}
CAV &= ACC(f^d,D)-ACC(f,D),
\end{aligned}
\end{equation}
where $ACC(f, D)$ denotes model $f$ accuracy on dataset $D$.

\emph{CRR/CSR.} %Classification Rectify or Sacrifice Ratio is designed to detail the difference of predictions before and after applying defenses to assess how they influence the predictions of models. 
CRR is the percentage of testing examples that are misclassified by $f$ previously but correctly classified by $f^d$. Inversely, CSR is the percentage of testing examples that are correctly classified by $f$ but
misclassified by $f^d$. Thus, they are defined as follows:
\begin{equation}
\begin{aligned}
CRR=\frac{1}{n}\sum_{i=1}^{n}{count((f(\mathbf{x}_i)\not=\mathbf{y}_i) \& (f^d(\mathbf{x}_i)=\mathbf{y}_i))},
\end{aligned}
\end{equation}
\begin{equation}
\begin{aligned}
CSR=\frac{1}{n}\sum_{i=1}^{n}{count((f(\mathbf{x}_i)=\mathbf{y}_i) \& (f^d(\mathbf{x}_i)\not=\mathbf{y}_i))},
\end{aligned}
\end{equation}
where $n$ is the number of examples.

\emph{CCV.} Defense strategies may not have negative influences on the accuracy performance, however, the prediction confidence of correctly classified examples may decrease. Classification Confidence Variance (CCV) can measure the confidence variance induced by robust models:
\begin{equation}
\begin{aligned}
CCV=\frac{1}{m}\sum_{i=1}^{m}{|P_{\mathbf{y}_i}(\mathbf{x}_i)-P^d_{\mathbf{y}_i}(\mathbf{x}_i)|},
\end{aligned}
\end{equation}
where $P_{\mathbf{y}_i}(\mathbf{x}_i)$ denotes the prediction confidence of model $f$ towards $\mathbf{y}_i$ and $m$ is the number of examples correctly classified by both $f$ and $f^d$.

\emph{COS.} Classification Output Stability (COS) uses JS divergence to measure the similarity of the classification output stability between the original model and the robust model as:
\begin{equation}
\begin{aligned}
COS=\frac{1}{m}\sum_{i=1}^{m}JSD({P(\mathbf{x}_i)\|P^d(\mathbf{x}_i)}),
\end{aligned}
\end{equation}
where $P(\mathbf{x}_i)$ and $P^d(\mathbf{x}_i)$ denotes the prediction confidence of model $f$ and $f^d$ on $\mathbf{x}_i$, respectively. $m$ is the number of examples correctly classified by both $f$ and $f^d$. {JSD stands for the JS divergence, which is a commonly used method of measuring the similarity between two probability distributions.}

\subsubsection{\textbf{Model Structures}} 

\begin{itemize}
\item \textbf{Boundary-based}
\end{itemize}

\emph{Empirical Boundary Distance (EBD)}. The minimum distance to the decision boundary among data points reflects the model robustness to small noises. EBD calculates the minimum distance to the model decision boundary in a heuristic way. A larger EBD value means a stronger model. Given a learnt model $f$ and point $\mathbf{x}_i$ with class label $\mathbf{y}_i$ ($i=1,\ldots,k$), it first generates a set $V$ of $m$ random orthogonal directions \cite{he2018decision}. Then, for each direction in $V$ it estimates the root mean square (RMS) distances $\phi_i(V)$ to the decision boundary of $f$, until the model's prediction changes, i.e., $f(\mathbf{x}_i) \neq \mathbf{y}_i$. Among $\phi_i(V)$, $d_i$ denotes the minimum distance moved to change the prediction for instance $\mathbf{x}_i$. Then, EBD is defined as follows:
\begin{equation}
\begin{aligned}
EBD=\frac{1}{n}\sum_{i=1}^{n} d_i, \quad d_i= \min\phi_i(V),
\end{aligned}
\end{equation}
where $n$ denotes the number of instances used.

\emph{Empirical Boundary Distance-2 (EBD-2)}. Additionally, we introduce the evaluation metrics EBD-2, which calculates the minimum distance of the model decision boundary for each class. Given a learnt model $f$ and dataset $\mathbb{D}=\{\mathbf x^{(i)}, \mathbf y^{(i)}\}_{i=1...n}$, for each direction $j$ in the $k$ classes, the metric estimates the distances $d_j$ to change the model prediction of $\mathbf x^{(i)}$, i.e., $f(\mathbf x^{(i)}) \neq \mathbf y^{(i)}$. Specifically, we use iterative adversarial attacks (e.g., BIM) in practice and calculate the steps used as the distance $d_j$. 

\begin{itemize}
\item \textbf{Consistency-based}
\end{itemize}

\emph{$\varepsilon$-Empirical Noise Insensitivity}. \cite{xu2012robustness} first introduced the concept of learning algorithms robustness from the idea that if two samples are ``similar'' then their test errors are very close. $\varepsilon$-Empirical Noise Insensitivity measures the model robustness against noise from the view of Lipschitz constant, and a lower value indicates a stronger model. We first select $n$ clean examples randomly, then $m$ examples are generated from each clean example via various methods, e.g., adversarial attack, Gaussian noise, blur, etc. The differences between model loss function are computed when clean example and corresponding polluted examples are fed to. The different severities in loss function is used to measure model insensitivity and stability to generalized small noise within constraint $\varepsilon$:
\begin{equation}
\begin{aligned}
I_f(\varepsilon)=&\frac{1}{n \times m}\sum_{i=1}^{n} \sum_{j=1}^{m} \frac{|l_f(\mathbf{x}_i|\mathbf{y}_{i})-l_f(\mu_{ij}|\mathbf{y}_{i})|}{|\mathbf{x}_i-\mu_{ij}|_\infty}\\
&s.t. \quad |\mathbf{x}_i-\mu_{ij}|_\infty \leq \varepsilon,
\end{aligned}
\end{equation}
where $\mathbf{x}_i$, $\mu_{ij}$ and $\mathbf{y}_i$ denote the clean example, corresponding polluted example and the class label, respectively. 

\begin{itemize}
\item \textbf{Neuron-based}
\end{itemize}

\emph{Neuron Sensitivity.} Intuitively, for a model that owns strong robustness, namely, insensitive to adversarial examples, the clean example $\mathbf{x}$ and the corresponding adversarial example $\mathbf{x}^\prime$ share a similar representation in the hidden layers of the model \cite{xu2012robustness}. Neuron Sensitivity can be deemed as the deviation of the feature representation in hidden layers between clean examples and corresponding adversarial examples, which measures model robustness from the perspective of neuron. Specifically, given a benign example $\mathbf{x}_i$, where $i=1,\ldots, \mathit{n}$, from $\mathbb{D}$ and its corresponding adversarial example $\mathbf{x}^\prime_i$ from $\mathbb{D}^{\prime}$, we can get the dual pair set $\bar{\mathbb{D}}=\{(\mathbf{x}_i,\mathbf{x}^{\prime}_i)\}$, and then calculate the neuron sensitivity $\sigma$ as follows:
\begin{equation}
\sigma(f_{l,m},\bar{\mathbb{D}})=\frac{1}{\mathit{n}}\sum_{i=1}^\mathit{n}\frac{1}{dim(f_{l,m}(\mathbf{x}_i))}\Vert f_{l,m}(\mathbf{x}_i)-f_{l,m}(\mathbf{x}_i^\prime)\Vert_1 ,\label{Neuron Sensitivity}
\end{equation}
where $f_{l,m}(\mathbf{x}_i)$ and $f_l^m(\mathbf{x}_i^\prime)$ respectively represents outputs of the $m$-th neuron at the $l$-th layer of $f$ towards clean example $\mathbf{x}_i$ and corresponding adversarial example $\mathbf{x}_i^\prime$ during the forward process. $dim(\cdot)$ denotes the dimension.

\emph{Neuron Uncertainty.} Model uncertainty has been widely investigated in safety critical applications to induce the confidence and uncertainty behaviors during model prediction. Motivated by the fact that model uncertainty is commonly induced by predictive variance, we use the variance of neuron $f_{l,m}$ to calculate the Neuron Uncertainty as:
\begin{equation}
\begin{aligned}
U(f_{l,m})=\frac{1}{n}\sum_{i=1}^{n}variance(f_{l,m}(\mathbf{x}^{(i)})).
\end{aligned}
\end{equation}

\section{Experiments}
\label{Section:exp}
Here, we evaluate model robustness using our proposed evaluation framework on image classification benchmarks CIFAR-10, SVHN, and ImageNet. {The experimental results of VGG-16 on SVHN and CIFAR-10 on WideResNet-28 can be found in supplementary materials.}

\begin{table}[!htb]
\vspace{-0.2in}
\centering
\caption{White-box adversarial attacks (\%) on CIFAR-10 using {ResNet-18} and ImageNet using ResNet-50.}
\vspace{-0.4in}
\begin{center}

\begin{sc}
\resizebox{\linewidth}{!}{
\subtable[{ResNet-18 on CIFAR-10}]{
\begin{tabular}{c|cccccc}
\toprule
& {Clean} & {PGD-$\ell_1$} & {PGD-$\ell_2$} & {PGD-$\ell_{\infty}$} & {AA} & {C\&W} \\ 
\hline
{Vanilla} & {89.4} & {51.5} & {0.9} & {0.0} & {0.0} & {0.4} \\
{PAT} & {78.7} & {71.7} & {52.3} & {35.1} & {34.5} & {52.1} \\
{TRADES} & {75.8} & {66.8} & {50.1} & {36.7} & {35.5} & {49.7} \\
{ALP} & {74.5} & {68.0} & {50.9} & {39.8} & {37.7} & {49.8} \\
{AWP} & {79.9} & {75.8} & {63.5} & {55.5} & {50.3} & {59.7} \\
\bottomrule
\end{tabular}
}
\subtable[{ResNet-50 on ImageNet}]{
\begin{tabular}{c|ccccc}
\toprule
& {Clean} & {PGD-$\ell_1$} & {PGD-$\ell_2$} & {PGD-$\ell_{\infty}$} & {AA} \\ 
\hline
{Vanilla} & {82.3} & {0.0} & {0.0} & {0.0} & {0.0} \\ 
{PAT} & {74.2} & {1.1} & {0.7} & {5.5} & {0.2} \\ 
{TRADES} & {40.6} & {7.9} & {0.3} & {4.4} & {0.1} \\ 
{ALP} & {56.2} & {1.2} & {0.1} & {0.4}& {0.0} \\ 
{Denoise} & {62.0} & {5.6} & {1.4} & {6.8} & {1.5} \\ 
\bottomrule
\end{tabular}
}
}
\end{sc}
\end{center}
\label{table:white-box}
\vspace{-0.2in}
\end{table}

\begin{table}[!htb]
\vspace{-0.2in}
\centering
\caption{Black-box adversarial attacks (\%) on CIFAR-10 using {ResNet-18} and ImageNet using ResNet-50.}
\vspace{-0.4in}
\begin{center}
\begin{sc}
\resizebox{\linewidth}{!}{

\subtable[{ResNet-18 on CIFAR-10}]{
\begin{tabular}{c|ccc|ccc}
\toprule
& {NAttack} & {SPSA} & {BA} & {PGD-$\ell_1$} & {PGD-$\ell_2$} & {PGD-$\ell_{\infty}$} \\ 
\hline
{Vanilla} & {0.0} & {8.8} & {0.0} & {81.2} & {53.0} & {20.9} \\
{PAT} & {35.8} & {71.9} & {0.0} & {78.6} & {78.2} & {77.1} \\
{TRADES} & {36.7} & {71.8} & {0.0} & {75.3} & {74.6} & {73.9} \\
{ALP} & {38.4} & {70.4} & {0.0} & {74.1} & {74.0} & {73.4} \\
{AWP} & {52.1} & {74.9} & {0.0} & {80.0} & {79.2} & {78.2} \\
\bottomrule
\end{tabular}
}
\subtable[{ResNet-50 on ImageNet}]{
\label{table:black-box-img}
\begin{tabular}{c|ccc|ccc}
\toprule
&{ NAttack} & {SPSA} & {BA} & {PGD-$\ell_1$} & {PGD-$\ell_2$} & {PGD-$\ell_{\infty}$} \\ 
\hline
{Vanilla} & {0.1} & {19.7} & {0.9} & {76.8} & {69.3} & {70.6} \\ 
{PAT} & {14.8} & {25.6} & {0.3} & {70.5} & {26.7} & {26.6} \\ 
{TRADES} & {8.3} & {42.7} & {1.3} & {39.3} & {37.6} & {37.8} \\ 
{ALP} & {21.6} & {34.7} & {0.2} & {58.6} & {56.4} & {55.7} \\ 
{Denoise} & {29.6} & {61.7} & {0.7} & {63.1} & {62.1} & {63.1} \\ 
\bottomrule
\end{tabular}
}
}
\end{sc}

\end{center}
\label{table:black-box}
\vspace{-0.4in}
\end{table}

\subsection{Model-oriented Evaluation}

\subsubsection{Model Behaviors}

As for adversarial robustness, we report metrics including CA, AAW, AAB, ACAC, ACTC, and NTE. The experimental results regarding CA and AAW can be found in Table \ref{table:white-box}; the results of AAB are shown in Table \ref{table:black-box}; and the results in terms of ACAC, ACTC, and NTE are listed in supplementary material. Besides standard black-box attacks (NAttack, SPSA, and BA), we also generate adversarial examples using an Inception-V3 then perform transfer attacks on the target model (denoted ``PGD-$\ell_1$'', ``PGD-$\ell_2$'', and ``PGD-$\ell_{\infty}$'' in Table \ref{table:black-box}).

As for corruption robustness,  the results of mCE, relative mCE, and mFR can be found in Figure \ref{fig:mCE}. Moreover, the results of CAV, CRR/CSR, CCV, and COS are illustrated in supplementary materials.

\begin{figure}[!htb]
\centering

\subfigure[{ResNet-18 on CIFAR-10}]{
\includegraphics[width=0.4\linewidth]{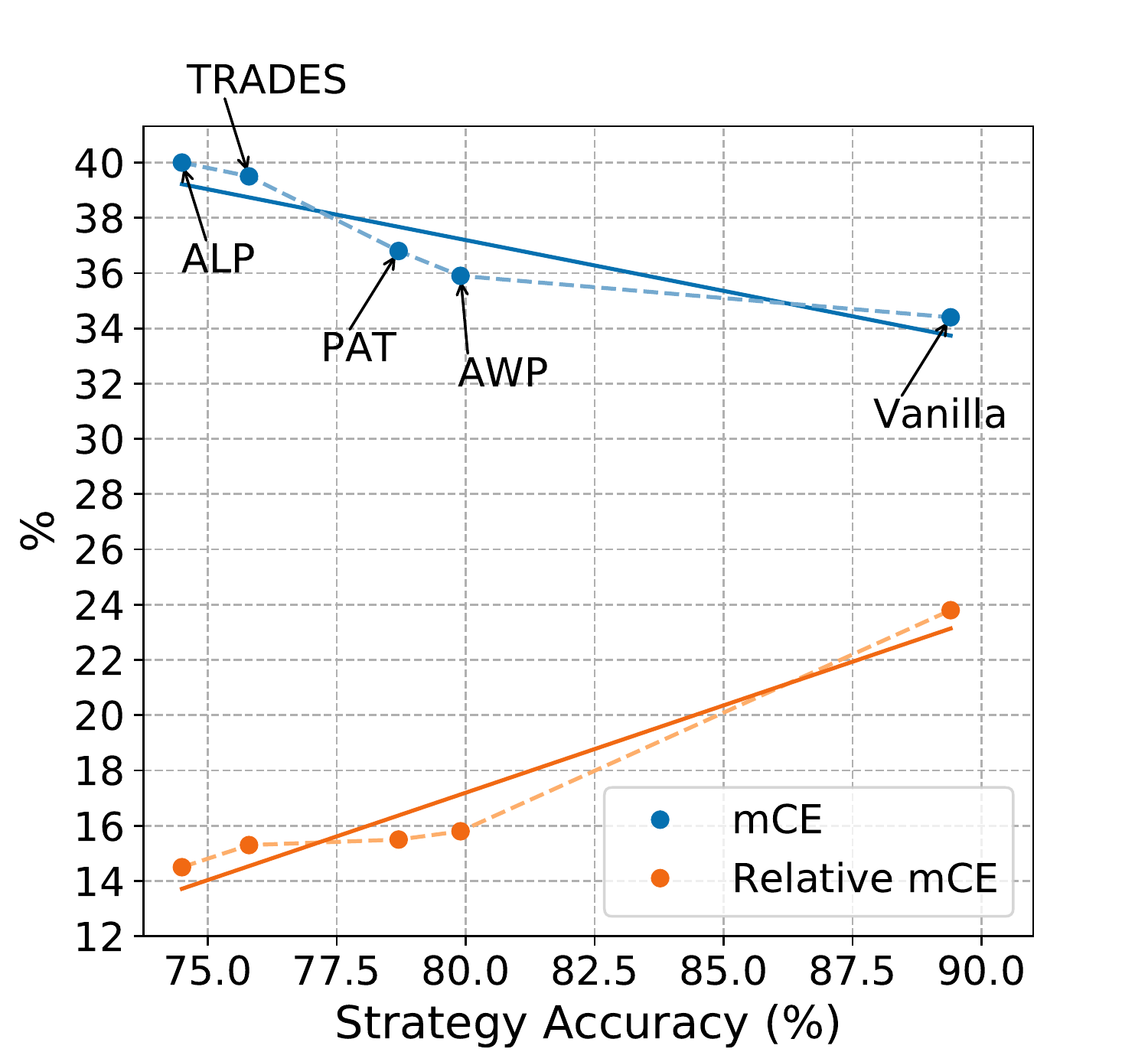}
}
\subfigure[{ResNet-50 on ImageNet}]{
\includegraphics[width=0.4\linewidth]{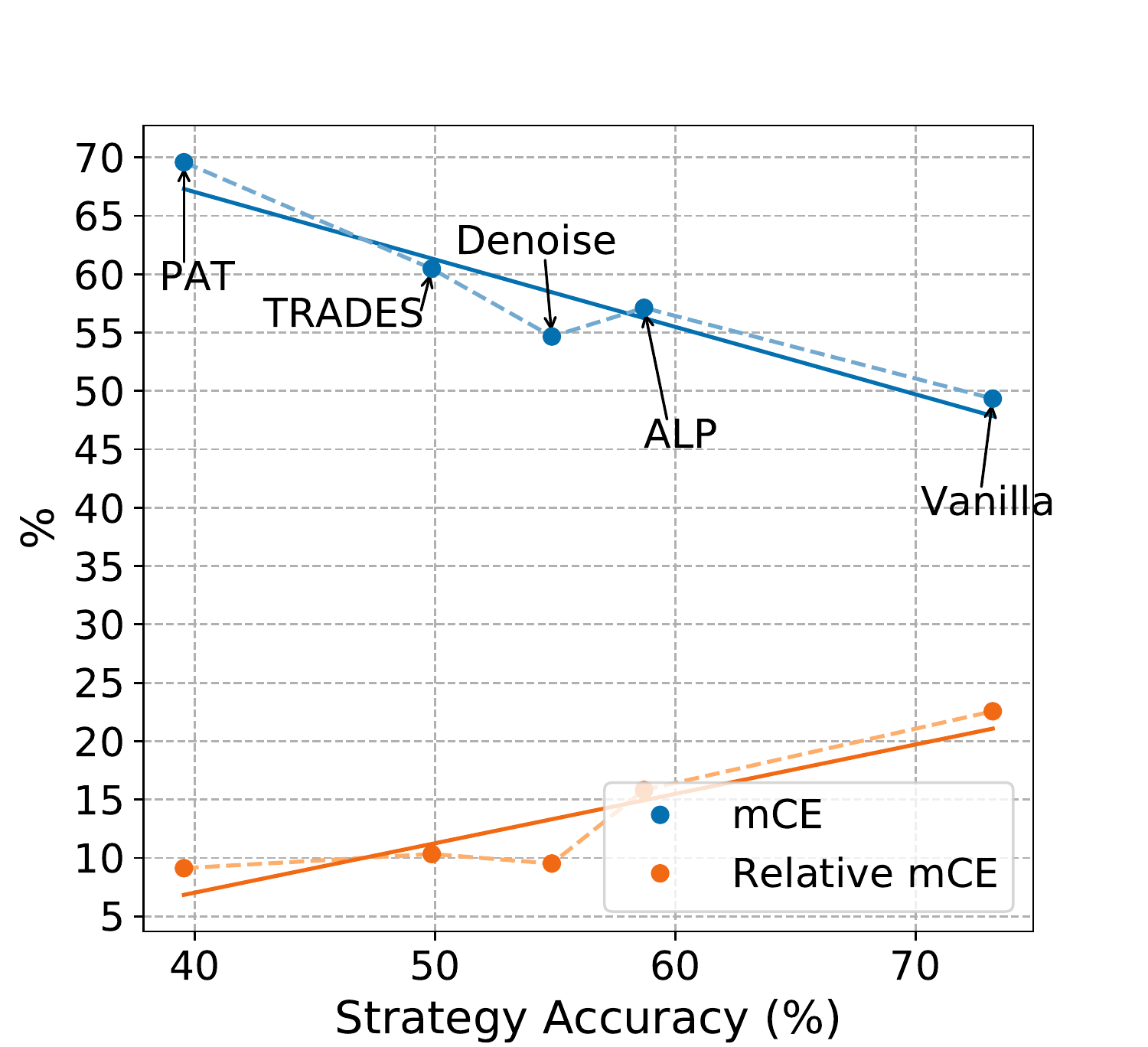}
}

\subfigure[{ResNet-18 on CIFAR-10}]{
\includegraphics[width=0.4\linewidth]{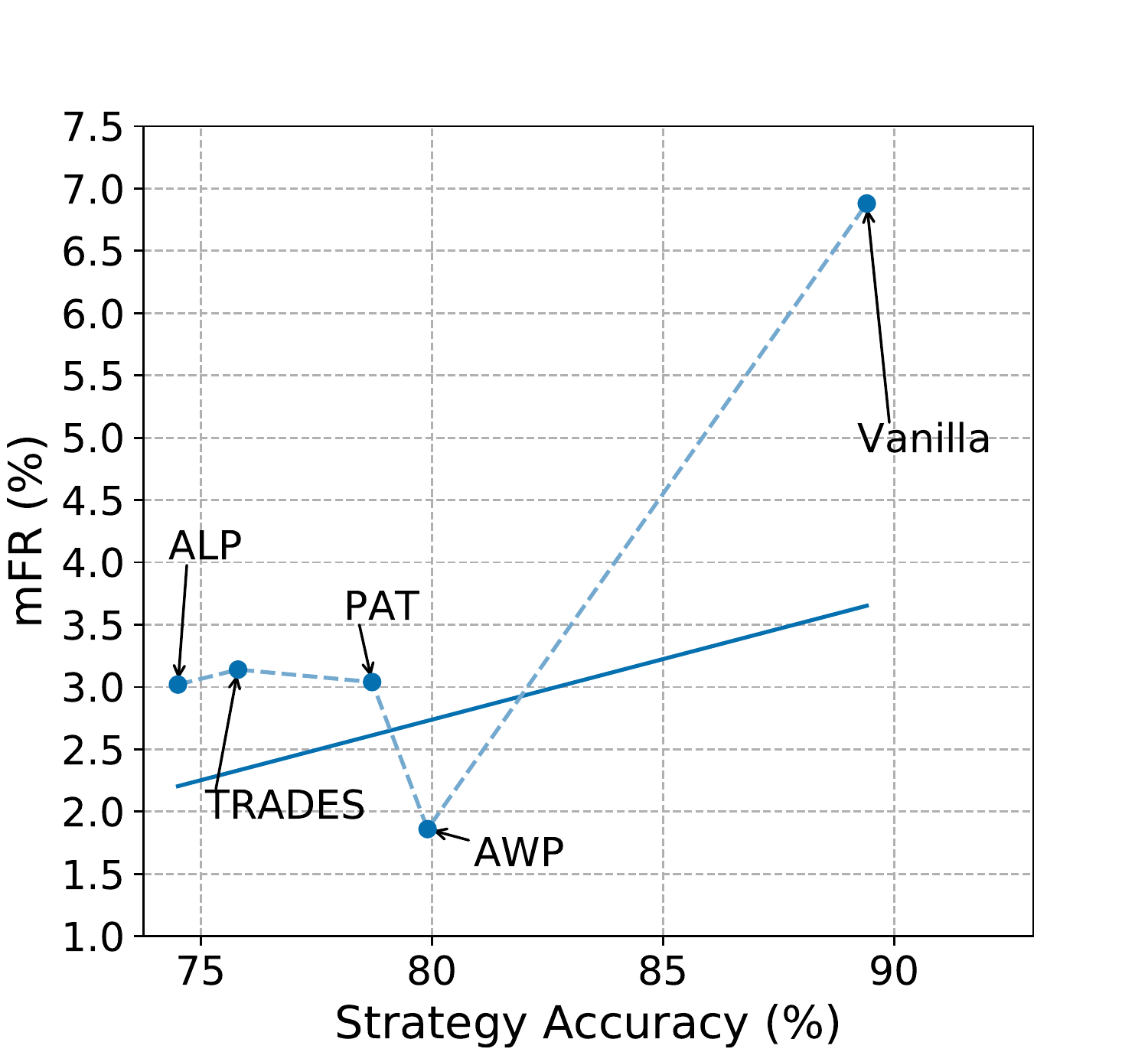}
}
\subfigure[{ResNet-50 on ImageNet}]{
\includegraphics[width=0.4\linewidth]{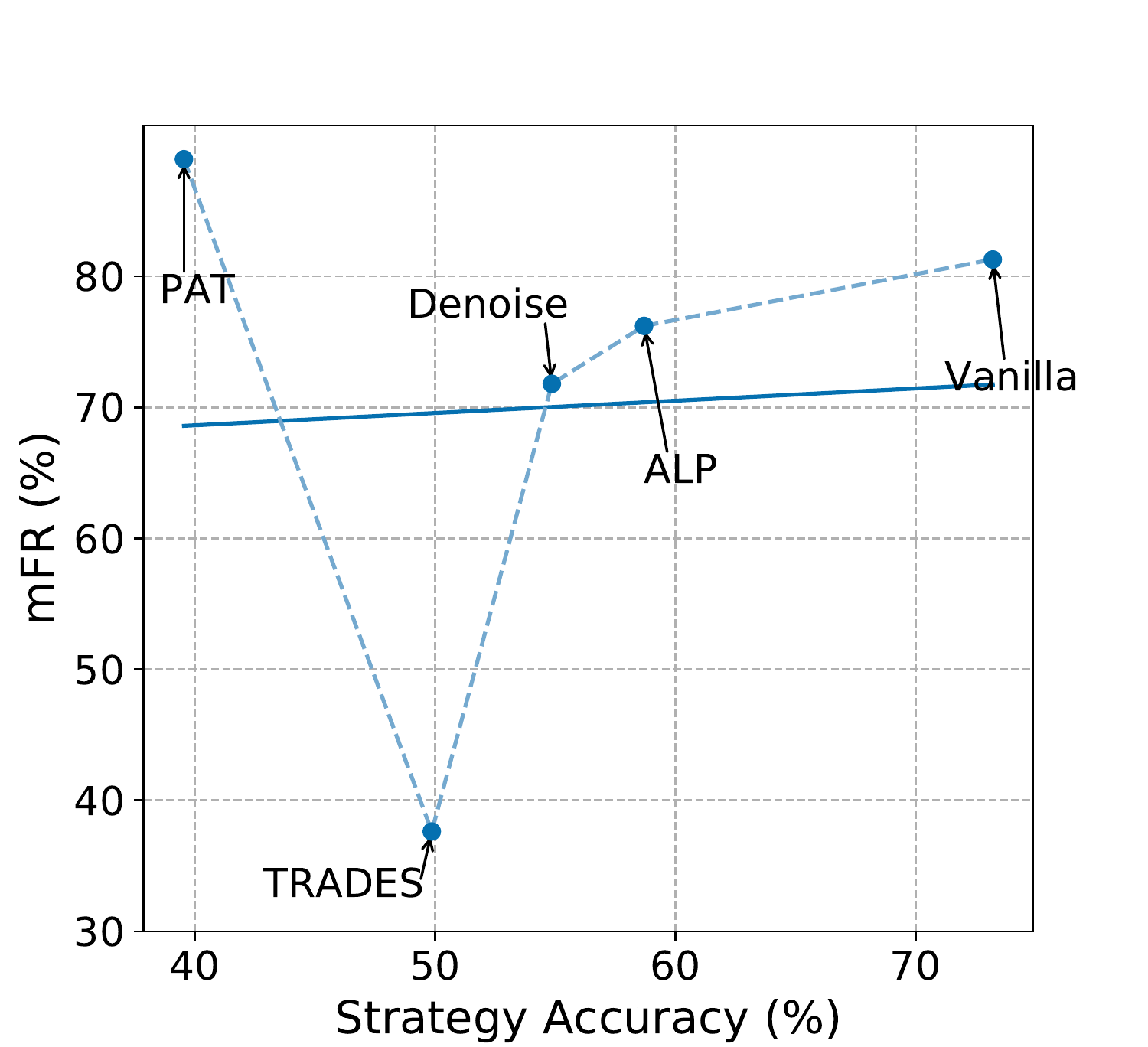}
}

\caption{Experimental results of mCE, RmCE, and mFR on CIFAR-10 and ImageNet corruption dataset.}
\label{fig:mCE}
\end{figure}

From the above experimental results, we can draw several conclusions as follows: (1) for small datasets like CIFAR-10 and SVHN, TRADES achieves the highest adversarial robustness for almost all adversarial attacks in both black-box and white-box settings, but it is vulnerable against corruptions; however, current defenses (especially adversarial training) are still suffering problems when scaling to large-scale datasets like ImageNet (e.g., TRADES fails to perform well on clean and adversarial attacks); (2) models trained on one specific perturbation type are vulnerable to other norm-bounded perturbations (e.g., $\ell_{\infty}$ trained models are weak towards $\ell_1$ and $\ell_2$ adversarial examples); (3) standard adversarially-trained models (SAT and PAT) are still vulnerable from a more rigorous perspective by showing high confidence of adversarial classes and low confidence of true classes; (4) transfer-based black-box attacks may fail to serve as a good indicator for showing robustness especially in large-scale dataset (see Table \ref{table:black-box-img}); and (5) as the model becomes increasingly robust, its COS and CCV values become comparatively high, which in turn indicates that the model is less stable towards clean examples. The reason might be there exists trade-off between standard accuracy and robust accuracy, thus models behaving high accuracy against adversarial attacks (i.e., TRADES) show low stability towards clean examples.

\subsubsection{Model Structures} 
\begin{figure}[!htb]
\centering

\subfigure[{ResNet-18 on CIFAR-10}]{
\includegraphics[width=0.4\linewidth]{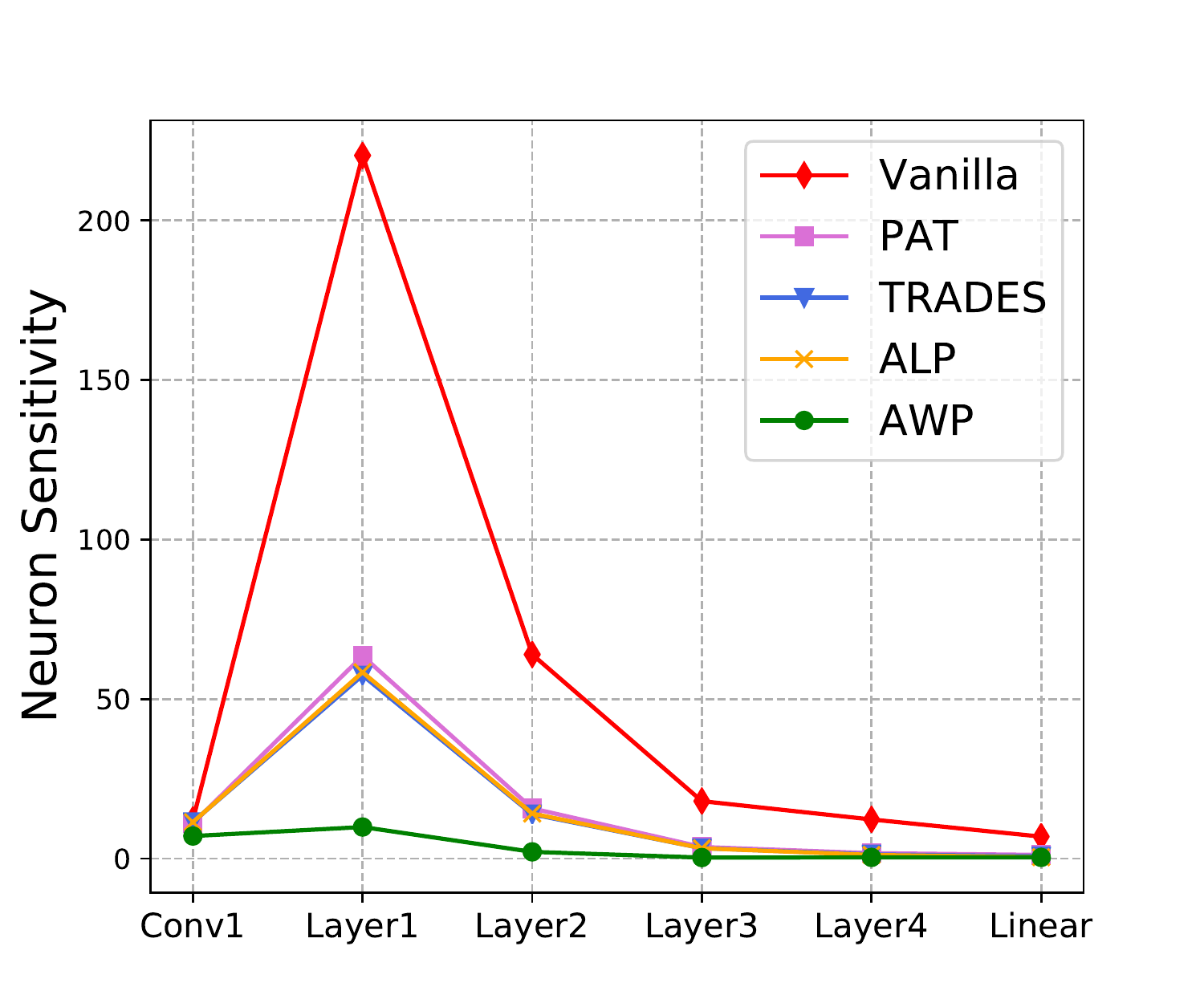}
}
\subfigure[{ResNet-50 on ImageNet}]{
\includegraphics[width=0.4\linewidth]{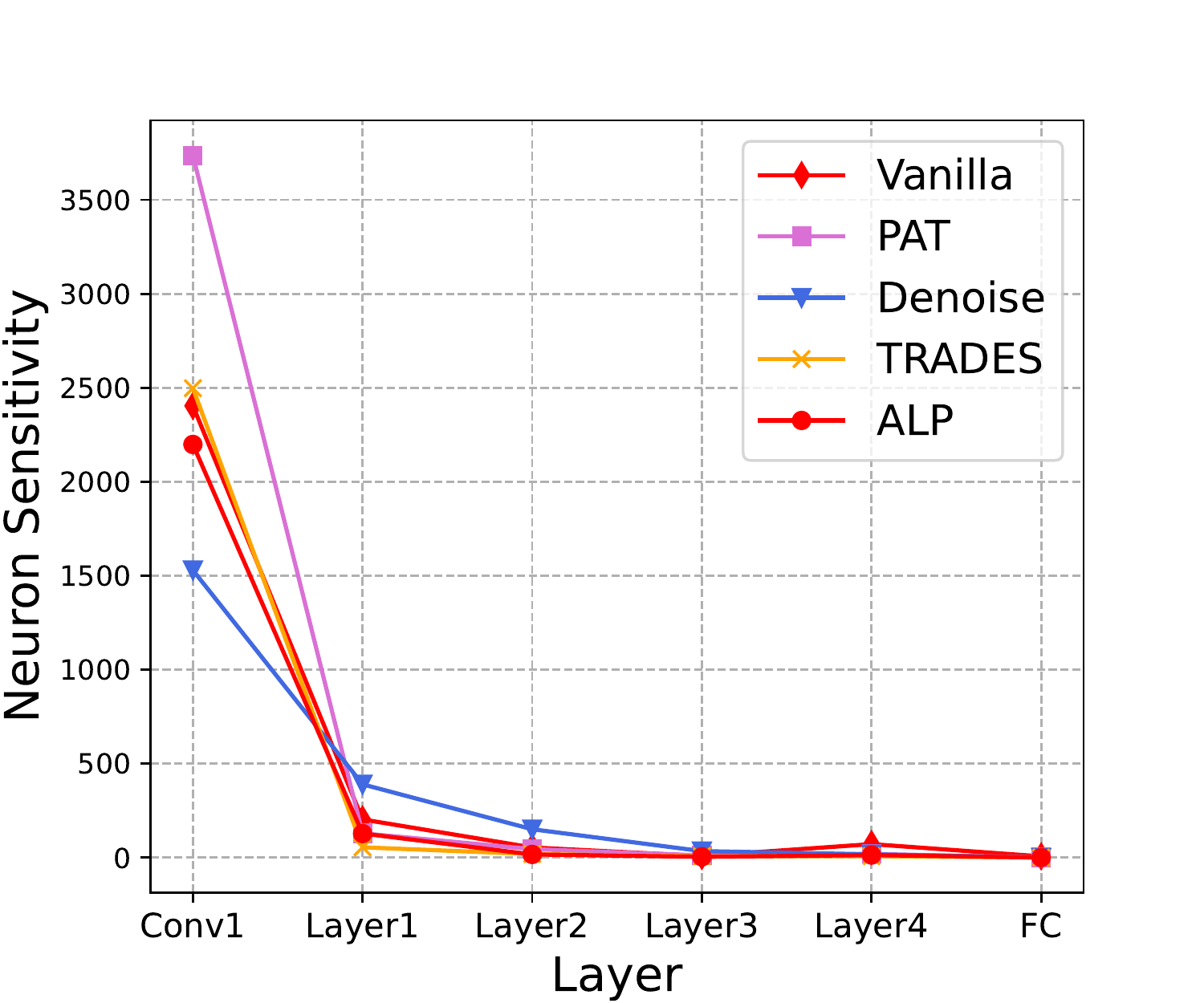}
}

\subfigure[{ResNet-18 on CIFAR-10}]{
\includegraphics[width=0.4\linewidth]{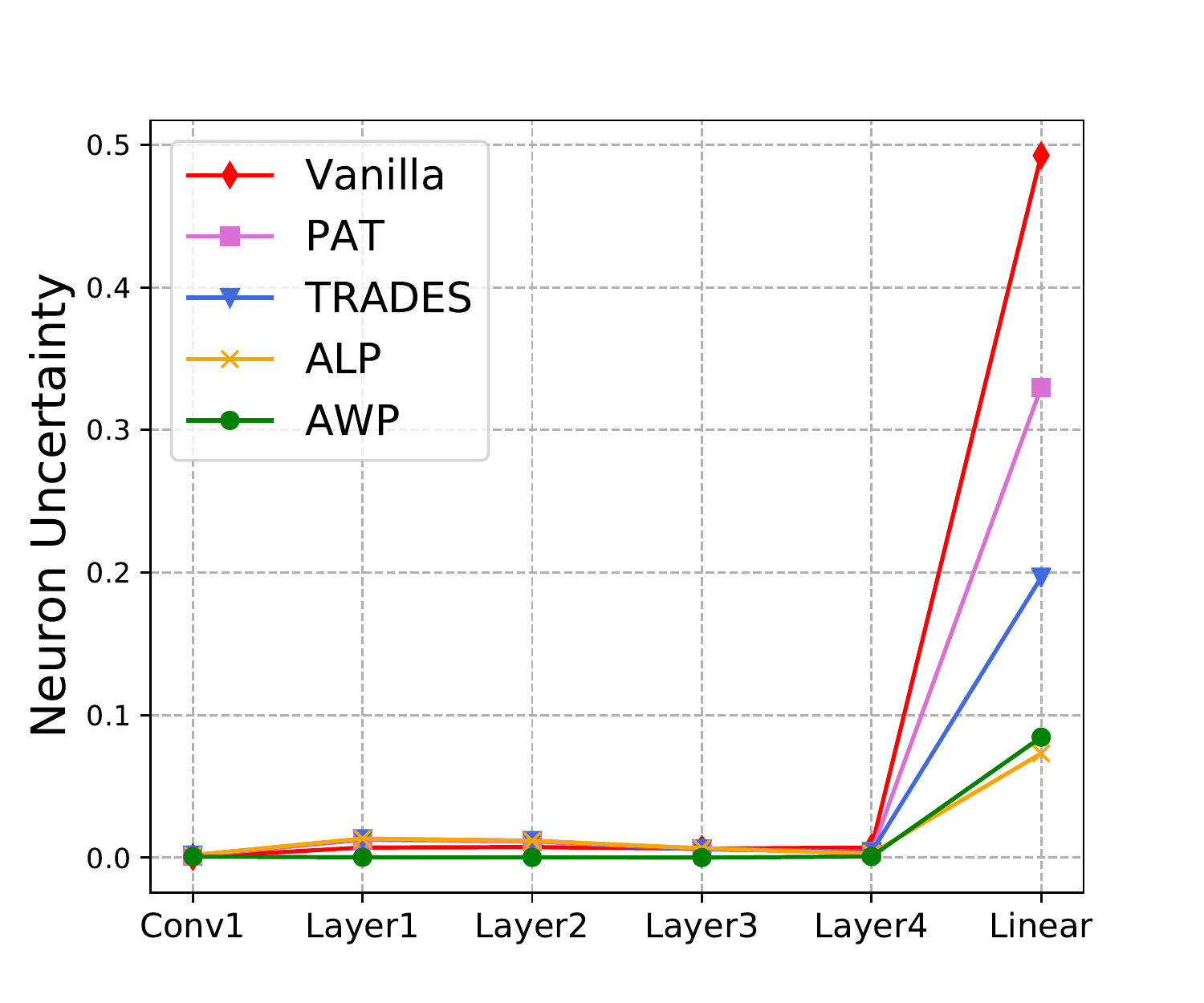}
}
\subfigure[{ResNet-50 on ImageNet}]{
\includegraphics[width=0.4\linewidth]{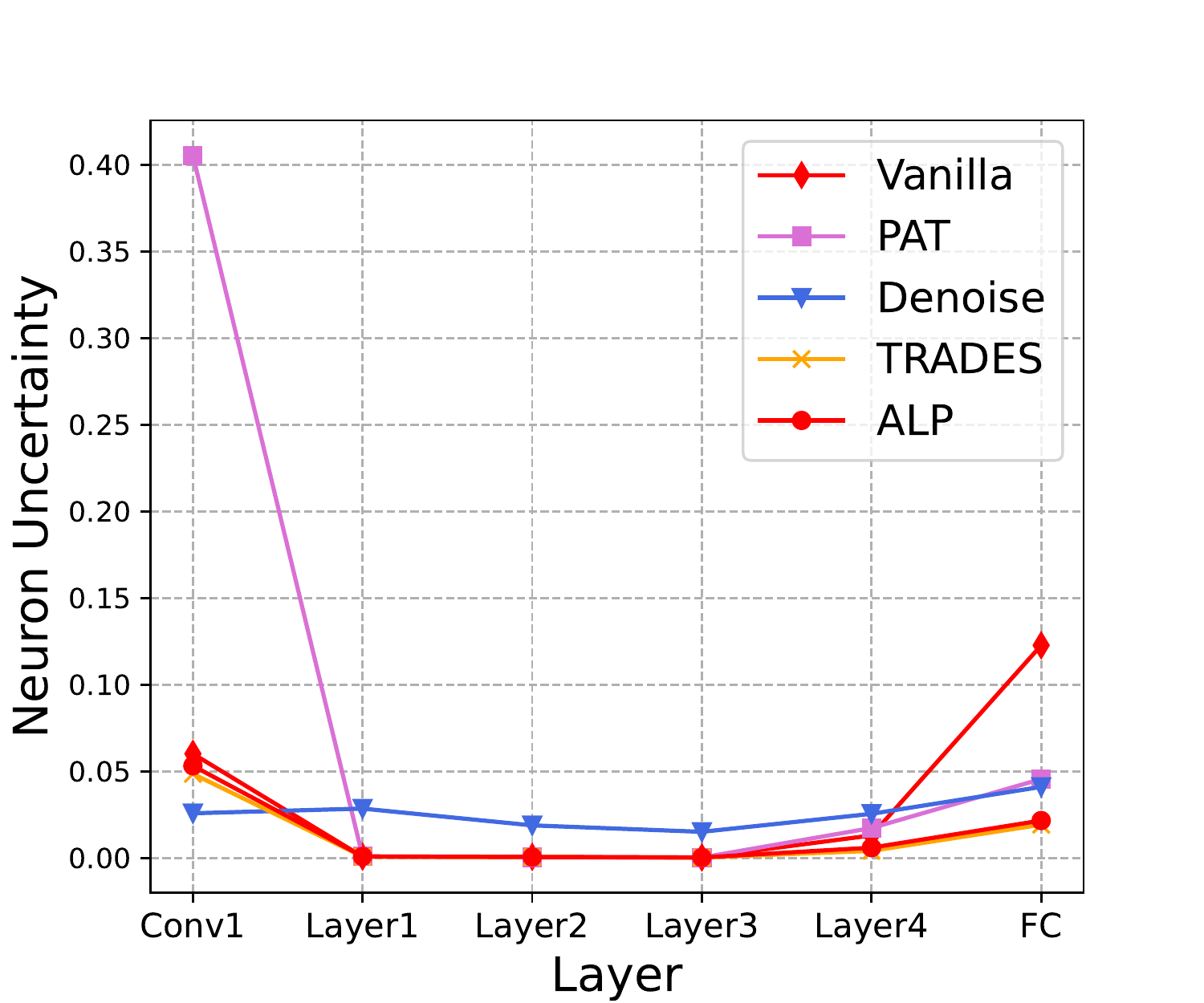}
}

%\vspace{-0.1in}
\caption{{Experimental results of Neuron Sensitivity and Neuron Uncertainty on CIFAR-10 using {ResNet-18} and ImageNet using ResNet-50, with PGD-$\ell_{\infty}$ adversarial examples. We report the mean value of the metrics for each layer.}}

\label{fig:neuronsen}
\end{figure}

We then evaluate model robustness with respect to structures. The results of EBD and EBD-2 are illustrated in Table \ref{table:EBD} and Fig \ref{fig:EBD}; the results of $\epsilon$-Empirical Noise Insensitivity can be found in supplementary material; Neuron Sensitivity and Neuron Uncertainty can be found in Figure \ref{fig:neuronsen}, respectively.

In summary, we can draw several interesting observations: \textit{(1) in most cases, models with higher adversarial accuracy are showing better structure robustness; (2) though showing the highest adversarial accuracy, TRADES does not have the largest EBD value as shown in Table \ref{table:EBD}.}

\begin{figure}[!htb]
\centering

\subfigure[{ResNet-18 on CIFAR-10}]{
\includegraphics[width=0.4\linewidth]{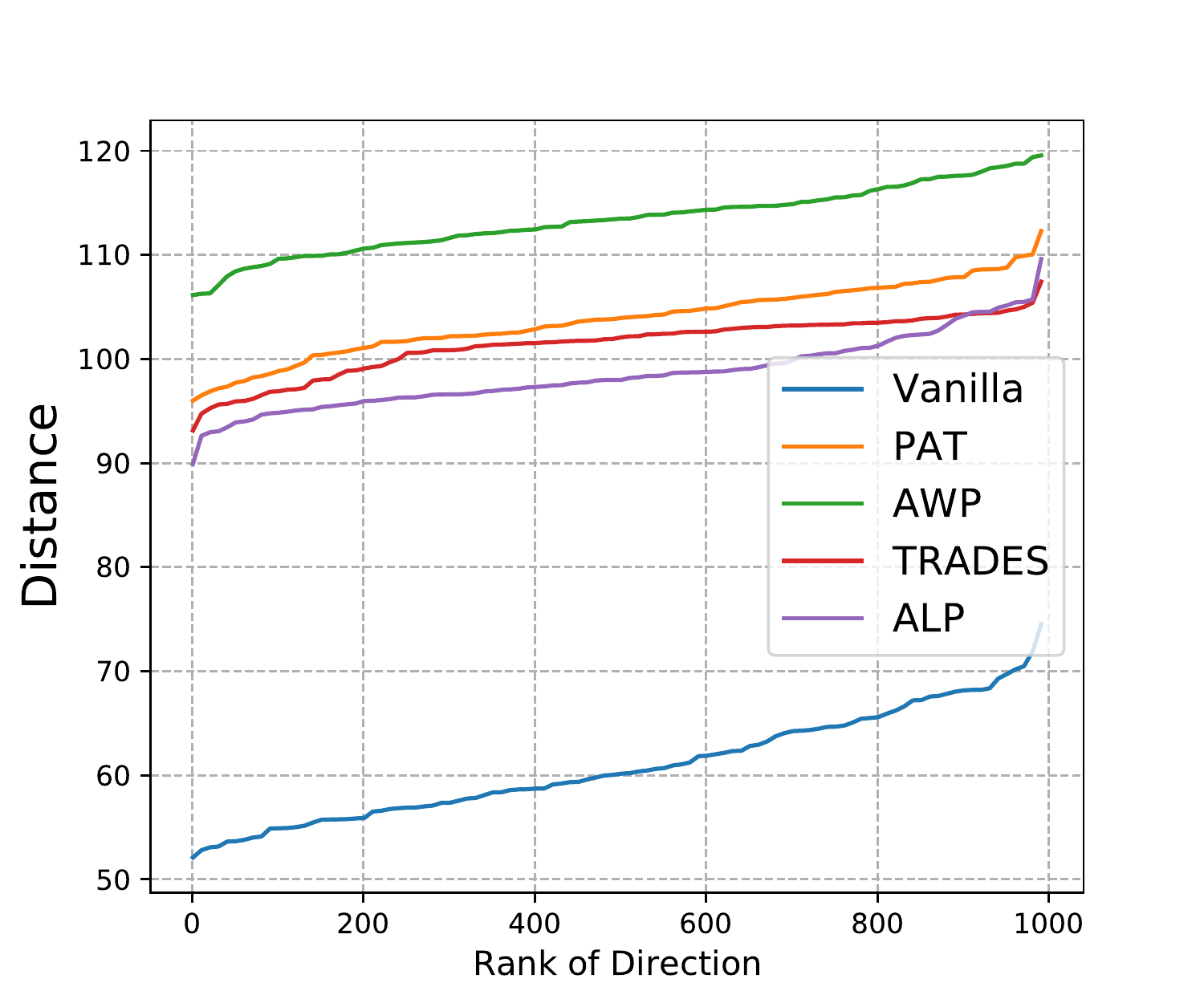}
}
\subfigure[{ResNet-50 on ImageNet}]{
\includegraphics[width=0.4\linewidth]{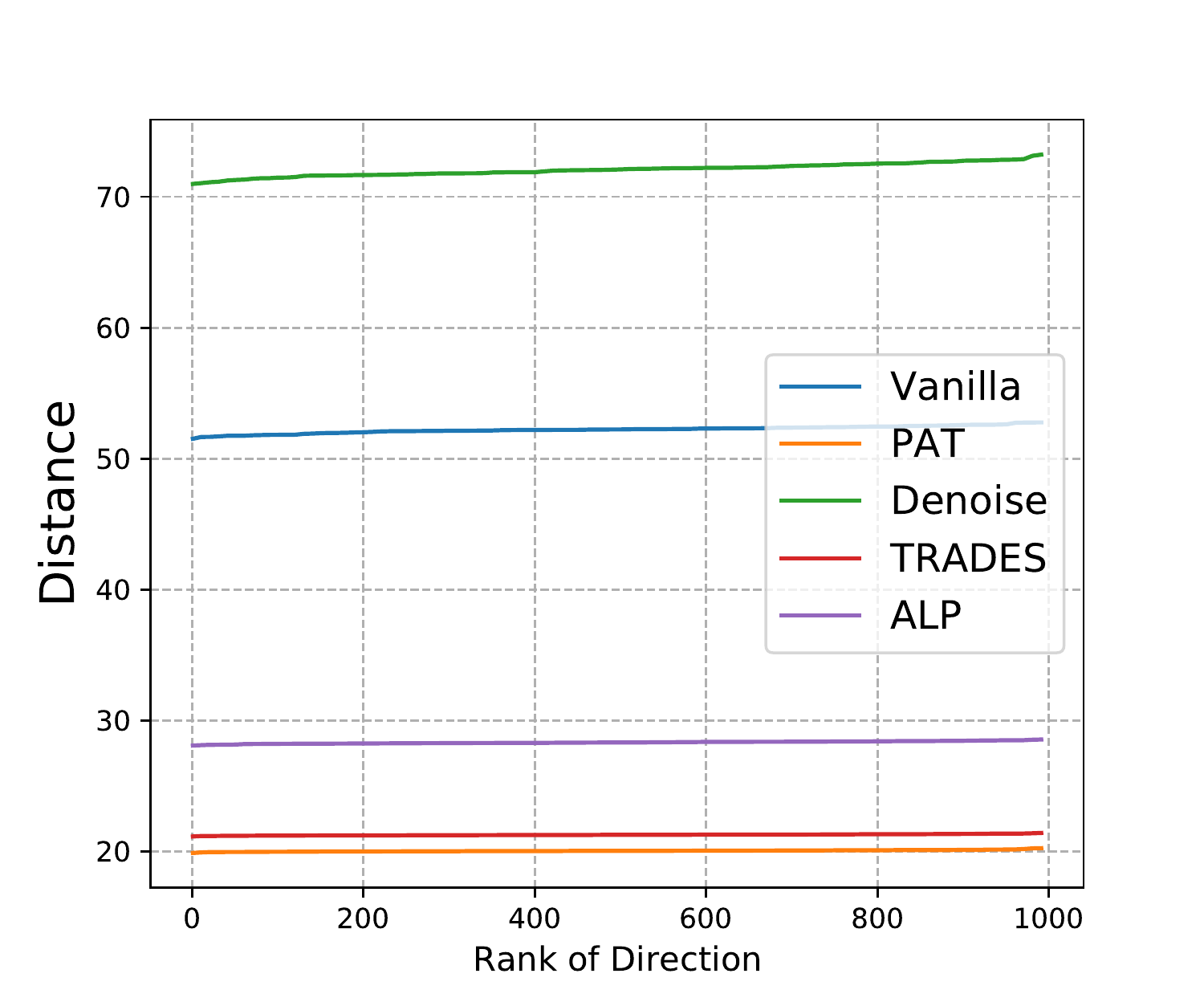}
}

\subfigure[{ResNet-18 on CIFAR-10}]{
\includegraphics[width=0.4\linewidth]{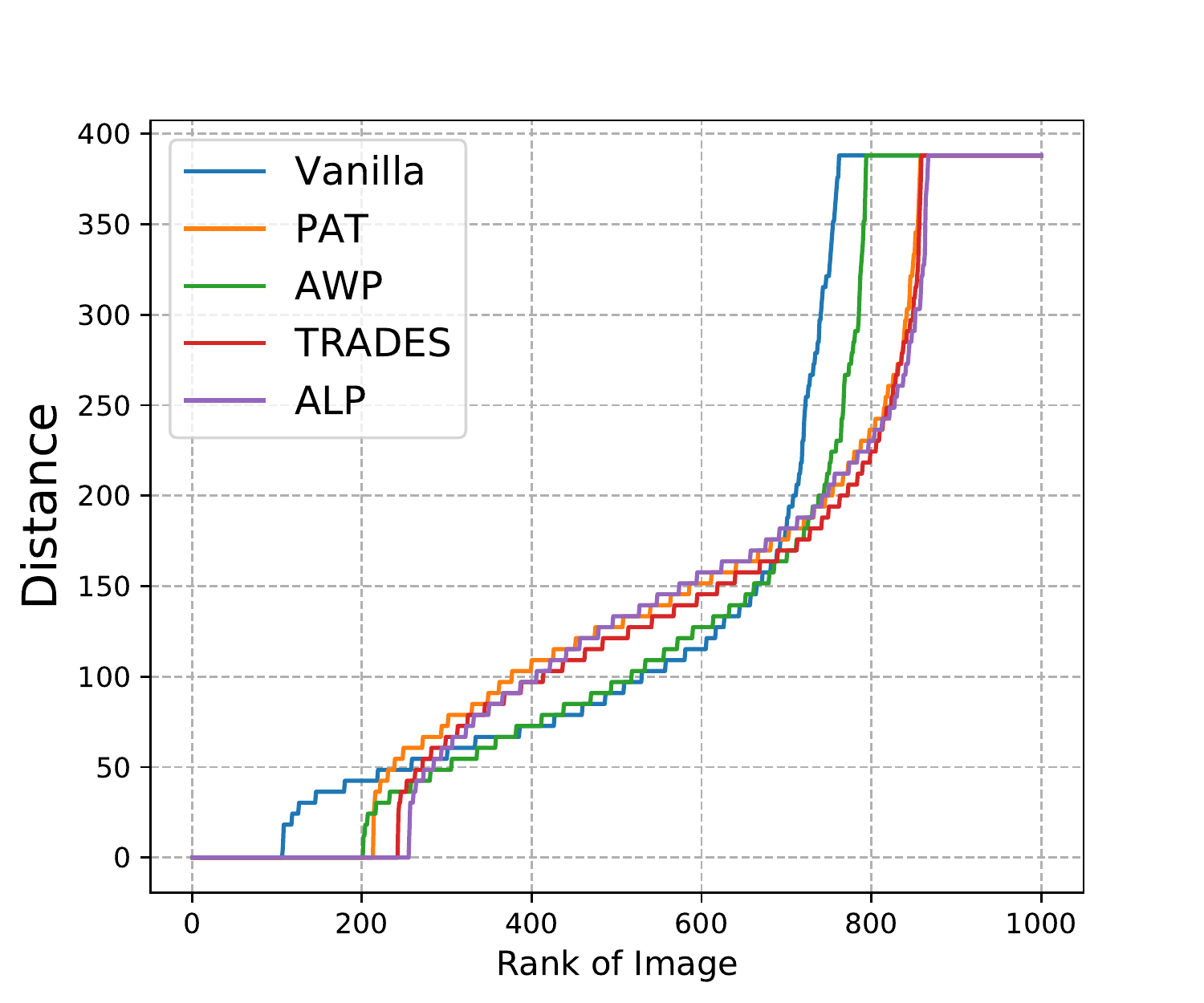}
}
\subfigure[{ResNet-50 on ImageNet}]{
\includegraphics[width=0.4\linewidth]{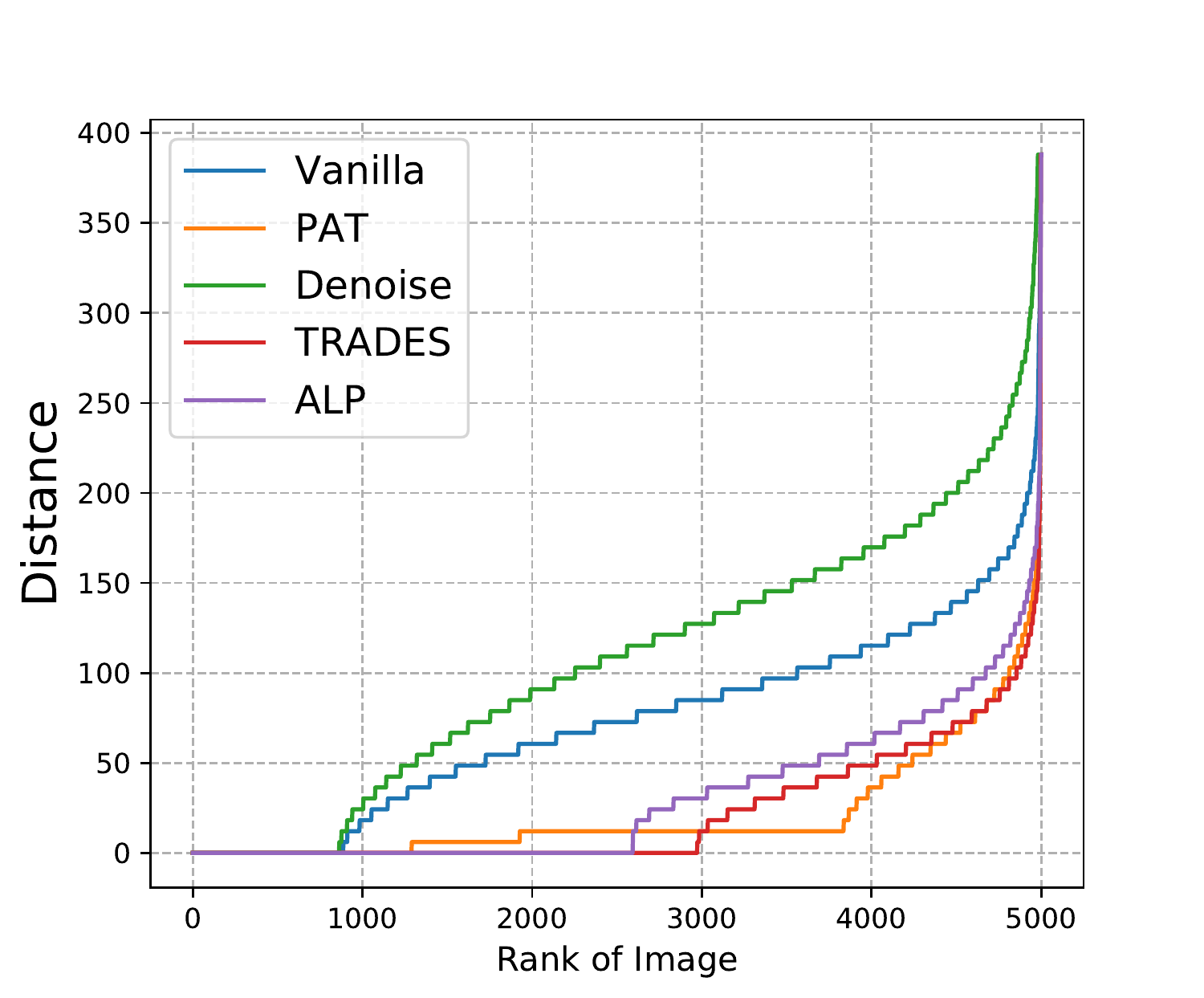}
}

%\vspace{-0.1in}
\caption{{Specific distance values on CIFAR10 and ImageNet related to EBD: (a)(b)(c) the average distance moved in each orthogonal direction, and (d)(e)(f) the Empirical Boundary Distance moved for 1000 different images.}}
\label{fig:EBD}
%\vspace{-0.1in}
\end{figure}

\begin{table}[!htb]
\vspace{-0.2in}
\centering
\caption{Experiments results of EBD (measured as RMS distance) and EBD-2 (measured as number of iterations with $\alpha=0.0005$) on CIFAR-10 using ResNet-18 and ImageNet using ResNet-50.}
\vspace{-0.2in}
\begin{center}

\begin{sc}
\resizebox{\linewidth}{!}{

\subtable[{ResNet-18 on CIFAR-10}]{
\begin{tabular}{c|cccccc}
\toprule
{Model} & {Vanilla} & {PAT} & {TRADES} & {ALP} & {AWP} \\
\hline
{EBD}     & {3.5} & {9.3} & {9.0} & {9.1} & {10.0} \\
{EBD-2}   & {8.6} & {46.9} & {46.4} & {48.3} & {61.0} \\
\bottomrule
\end{tabular}
}

\subtable[{ResNet-50 on ImageNet}]{
\begin{tabular}{c|cccccc}
\toprule
{Model} & {Vanilla} & {PAT} & {SAT} & {TRADES} & {ALP}  \\
\hline
{EBD}     & {35.2} & {18.1} & {20.0} & {26.1} & {45.2} \\
{EBD-2}   & {2.2} & {3.9} & {19.7} & {19.9} & {3.0} \\
\bottomrule
\end{tabular}
}

}

\end{sc}

\end{center}
\label{table:EBD}
\vspace{-0.2in}
\end{table}

\begin{table}[!htb]
\centering
\vspace{-0.2in}
\caption{Experiments results of KMNCov on CIFAR-10 using {ResNet-18} and ImageNet using ResNet-50.}
\vspace{-0.4in}
\begin{center}

\begin{sc}
\resizebox{\linewidth}{!}{
\subtable[{ResNet-18 on CIFAR-10}]{
\begin{tabular}{c|cccc|cc}
\toprule
& {FGSM} & {PGD-$\ell_1$} & {PGD-$\ell_2$} & {PGD-$\ell_{\infty}$} & {NAttack} & {SPSA} \\ 
\hline
{Vanilla} & {89.1} & {90.2} & {92.0} & {93.2} & {88.9} & {88.3} \\
{PAT} & {90.5} & {91.0} & {90.6} & {90.3} & {90.3} & {90.7} \\
{TRADES} & {91.1} & {91.2} & {91.0} & {90.9} & {90.9} & {91.0} \\
{ALP} & {89.7} & {90.9} & {90.0} & {89.6} & {89.7} & {90.5} \\
{AWP} & {83.0} & {87.0} & {84.1} & {83.2} & {83.6} & {86.2} \\
\bottomrule
\end{tabular}
}

\subtable[{ResNet-50 on ImageNet}]{
\begin{tabular}{c|cccc|cc}
\toprule
& {FGSM} & {PGD-$\ell_1$} & {PGD-$\ell_2$} & {PGD-$\ell_{\infty}$} & {NAttack} & {SPSA} \\ 
\hline
{Vanilla} & {56.7} & {82.6} & {85.9} & {86.6} & {58.5} & {59.3} \\ 
{PAT} & {47.0} & {69.2} & {47.7} & {47.4} & {52.5} & {47.2} \\ 
{TRADES} & {57.0} & {60.2} & {68.1} & {66.7} & {57.7} & {57.8} \\ 
{ALP} & {55.3} & {64.0} & {68.0} & {66.1} & {55.4} & {55.9} \\ 
{Denoise} & {58.9} & {60.5} & {64.2} & {59.7} & {33.0} & {59.2} \\ 
\bottomrule
\end{tabular}
}

}
\end{sc}

\end{center}
\label{table:KMNCov}
\vspace{-0.2in}
\end{table}

\begin{table}[!htb]
\centering
\vspace{-0.2in}
\caption{Experiments results of NBCov on CIFAR-10 using {ResNet-18} and ImageNet using ResNet-50.}
\vspace{-0.4in}
\begin{center}

\begin{sc}
\resizebox{\linewidth}{!}{

\subtable[{ResNet-18 on CIFAR-10}]{
\begin{tabular}{c|cccc|cc}
\toprule
& {FGSM} & {PGD-$\ell_1$} & {PGD-$\ell_2$} & {PGD-$\ell_{\infty}$} & {NAttack} & {SPSA} \\ 
\hline
{Vanilla} & {9.4} & {7.1} & {15.4} & {27.8} & {6.9} & {7.7} \\
{PAT} & {7.5} & {7.0} & {7.0} & {7.2} & {7.1} & {7.0} \\
{TRADES} & {7.7} & {7.3} & {7.3} & {7.4} & {7.4} & {6.9} \\
{ALP} & {6.9} & {6.8} & {6.8} & {6.8} & {6.9} & {6.5} \\
{AWP} & {6.1} & {6.2} & {6.2} & {6.2} & {6.2} & {6.1} \\
\bottomrule
\end{tabular}
}

\subtable[{ResNet-50 on ImageNet}]{
\begin{tabular}{c|cccc|cc}
\toprule
& {FGSM} & {PGD-$\ell_1$} & {PGD-$\ell_2$} & {PGD-$\ell_{\infty}$} & {NAttack} & {SPSA} \\ 
\hline
{Vanilla} & {58.8} & {91.3} & {88.4} & {90.6} & {57.0} & {60.6} \\ 
{PAT} & {54.9} & {92.8} & {56.2} & {57.4} & {58.3} & {54.5} \\ 
{TRADES} & {57.7} & {81.4} & {84.4} & {83.0} & {52.6} & {55.2} \\ 
{ALP} & {65.2} & {90.2} & {88.4} & {90.1} & {55.1} & {62.0} \\ 
{Denoise} & {45.7} & {66.7} & {82.1} & {58.8} & {3.6} & {47.8} \\ 
\bottomrule
\end{tabular}
}
}
\end{sc}

\end{center}
\label{table:NBCov}
\vspace{-0.2in}
\end{table}

\begin{table}[!htb]
\centering
\vspace{-0.2in}
\caption{Experiments results of SNACov on CIFAR-10 using {ResNet-18} and ImageNet using ResNet-50.}
\vspace{-0.4in}
\begin{center}

\begin{sc}
\resizebox{\linewidth}{!}{

\subtable[{ResNet-18 on CIFAR-10}]{
\begin{tabular}{c|cccc|cc}
\toprule
& {FGSM} & {PGD-$\ell_1$} & {PGD-$\ell_2$} & {PGD-$\ell_{\infty}$} & {NAttack} & {SPSA} \\ 
\hline
{Vanilla} & {4.7} & {2.2} & {12.6} & {27.3} & {2.0} & {3.0} \\
{PAT} & {2.8} & {2.2} & {2.3} & {2.5} & {2.2} & {2.3} \\
{TRADES} & {3.1} & {2.5} & {2.5} & {2.7} & {2.4} & {2.2} \\
{ALP} & {2.3} & {2.0} & {2.1} & {2.1} & {1.9} & {1.9} \\
{AWP} & {1.0} & {1.0} & {1.2} & {1.1} & {1.0} & {1.1} \\
\bottomrule
\end{tabular}
}

\subtable[{ResNet-50 on ImageNet}]{
\begin{tabular}{c|cccc|cc}
\toprule
& {FGSM} & {PGD-$\ell_1$} & {PGD-$\ell_2$} & {PGD-$\ell_{\infty}$} & {NAttack} & {SPSA} \\ 
\hline
{Vanilla} & {29.3} & {83.0} & {78.1} & {82.0} & {23.8} & {30.7} \\ 
{PAT} & {27.4} & {86.6} & {29.7} & {30.2} & {28.2} & {26.9} \\ 
{TRADES} & {27.7} & {68.6} & {76.0} & {72.0} & {10.2} & {24.6} \\ 
{ALP} & {40.0} & {81.8} & {78.7} & {81.7} & {17.2} & {33.1} \\ 
{Denoise} & {41.5} & {61.6} & {75.1} & {55.1} & {1.4} & {47.1} \\ 
\bottomrule
\end{tabular}
}
}
\end{sc}

\end{center}
\label{table:SNACov}
\vspace{-0.2in}
\end{table}

\subsection{Data-oriented Evaluation}
We then report the data-oriented evaluation metrics. {We randomly sample 1000 images from test set for CIFAR-10 and SVHN, and 5000 images for ImageNet; we then adversarially perturb these images using FGSM and PGD, respectively.} We finally compute and report the neuron-coverage related metrics (KMNCov, NBCov, SNACov) using these test sets. The results can be found in Table \ref{table:KMNCov}, \ref{table:NBCov} and \ref{table:SNACov}. Further, we show the results of ALD$_p$, ASS, and PSD on these test sets in Table \ref{table:ALDp}, \ref{table:ASS} and \ref{table:PSD}.

\begin{table}[!htb]
\centering
\vspace{-0.2in}
\caption{Experiments results of ALD$_p$(\%) on CIFAR-10 using {ResNet-18} and ImageNet using ResNet-50.}
\vspace{-0.4in}
\begin{center}

% \begin{sc}
\resizebox{\linewidth}{!}{

\subtable[{ResNet-18 on CIFAR-10}]{
\begin{tabular}{c|cccc}
\toprule
& FGSM & {PGD-$\ell_1$} & {PGD-$\ell_2$} & {PGD-$\ell_{\infty}$} \\ 
\hline
{Vanilla} & {6.5} & {0.7} & {1.9} & {5.4} \\
{PAT} & {6.6} & {0.7} & {1.9} & {6.1} \\
{TRADES} & {6.5} & {0.7} & {1.9} & {6.1} \\
{ALP} & {6.6} & {0.7} & {1.9} & {6.1} \\
{AWP} & {6.5} & {0.7} & {1.9} & {6.0} \\
\bottomrule
\end{tabular}
}
\subtable[{ResNet-50 on ImageNet}]{
\begin{tabular}{c|cccc}
\toprule
& FGSM & {PGD-$\ell_1$} & {PGD-$\ell_2$} & {PGD-$\ell_{\infty}$}\\ 
\hline
{Vanilla} & {6.2} & {2.0} & {8.8} & {7.5} \\ 
{PAT} & {6.2} & {2.2} & {8.8} & {10.0} \\ 
{TRADES} & {6.2} & {3.8} & {8.8} & {10.1} \\ 
{ALP} & {6.3} & {3.7} & {8.8} & {9.0} \\ 
{Denoise} & {6.3} & {3.8} & {8.8} & {10.2} \\ 
\bottomrule
\end{tabular}
}

}

% \end{sc}

\end{center}
\label{table:ALDp}
\vspace{-0.2in}
\end{table}

\begin{table}[!htb]
\centering
\vspace{-0.2in}
\caption{Experiments results of ASS(\%) on CIFAR-10 using {ResNet-18} and ImageNet using ResNet-50.}
\vspace{-0.4in}
\begin{center}

\begin{sc}
\resizebox{\linewidth}{!}{

\subtable[{ResNet-18 on CIFAR-10}]{
\begin{tabular}{c|cccc}
\toprule
& FGSM & PGD-$\ell_1$ & PGD-$\ell_2$ & PGD-$\ell_{\infty}$ \\ 
\hline
{Vanilla} & {93.0} & {99.9} & {99.1} & {95.2} \\
{PAT} & {95.5} & {100.0} & {99.7} & {96.1} \\
{TRADES} & {95.5} & {100.0} & {99.8} & {96.2} \\
{ALP} & {95.4} & {100.0} & {99.8} & {96.1} \\
{AWP} & {95.4} & {100.0} & {99.7} & {96.1} \\
\bottomrule
\end{tabular}
}
\subtable[{ResNet-50 on ImageNet}]{
\begin{tabular}{c|cccc}
\toprule
& {FGSM} & {PGD-$\ell_1$} & {PGD-$\ell_2$} & {PGD-$\ell_{\infty}$} \\ 
\hline
{Vanilla} & {82.1} & {97.3} & {77.9} & {77.5} \\ 
{PAT} & {82.0} & {97.4} & {83.0} & {82.0} \\ 
{TRADES} & {85.5} & {97.4} & {82.3} & {77.4} \\ 
{ALP} & {82.7} & {97.1} & {79.6} & {75.9} \\ 
{Denoise} & {87.3} & {98.0} & {81.3} & {78.6} \\ 
\bottomrule
\end{tabular}
}
}
\end{sc}

\end{center}
\label{table:ASS}
\vspace{-0.2in}
\end{table}

\begin{table}[!htb]
\centering
\vspace{-0.2in}
\caption{Experiments results of PSD(\%) on CIFAR-10 using {ResNet-18} and ImageNet using ResNet-50.}
\vspace{-0.4in}
\begin{center}

\begin{sc}
\resizebox{\linewidth}{!}{

\subtable[{ResNet-18 on CIFAR-10}]{
\begin{tabular}{c|cccc}
\toprule
& FGSM & PGD-$\ell_1$ & PGD-$\ell_2$ & PGD-$\ell_{\infty}$ \\ 
\hline
{Vanilla} & {70.9} & {2.9} & {22.7} & {59.1} \\
{PAT} & {159.5} & {1.1} & {11.7} & {153.8} \\
{TRADES} & {153.5} & {1.2} & {11.6} & {139.9} \\
{ALP} & {176.0} & {1.2} & {11.8} & {155.6} \\
{AWP} & {201.4} & {1.2} & {10.1} & {175.8} \\
\bottomrule
\end{tabular}
}
\subtable[{ResNet-50 on ImageNet}]{
\begin{tabular}{c|cccc}
\toprule
& {FGSM} & {PGD-$\ell_1$} & {PGD-$\ell_2$} & {PGD-$\ell_{\infty}$} \\ 
\hline
{Vanilla} & {127.0} & {17.0} & {90.0} & {69.4} \\ 
{PAT} & {171.0} & {16.6} & {88.0} & {227.1} \\ 
{TRADES} & {266.5} & {4.9} & {86.6} & {186.5} \\ 
{ALP} & {109.2} & {7.6} & {91.0} & {106.6} \\ 
{Denoise} & {326.0} & {4.4 }& {86.5} & {226.4} \\ 
\bottomrule
\end{tabular}
}
}
\end{sc}

\end{center}
\label{table:PSD}
\vspace{-0.4in}
\end{table}

In summary, we can draw conclusions as follows: \textit{(1) adversarial examples generated by $\ell_{\infty}$-norm attacks show significantly higher neuron coverage than other perturbation types (\eg, $\ell_1$ and $\ell_2$), which indicate that $\ell_{\infty}$-norm attacks cover more ``paths'' for a DNN when perform test or evaluation; (2) meanwhile, $\ell_{\infty}$-norm attacks are more imperceptible to the human vision (lower ALD$_p$, PSD, and higher ASS values compared to $\ell_1$ and $\ell_2$ attacks).}

\section{Discussions and Suggestions}
\label{Section:discussion}
Having demonstrated extensive experiments on these datasets using our comprehensive evaluation framework, we now take a further step and provide additional suggestions to the evaluation of model robustness as well as the design of adversarial attacks/defenses in the future.

\subsection{Evaluate Model Robustness using More Attacks}
For most studies in the adversarial learning literature\cite{liu2019training,zhang2019interpreting,xie2018mitigating}, they evaluate model robustness primarily on $\ell_{\infty}$-norm bounded PGD attacks, which has been shown to be the most effective and representative adversarial attack. However, according to our experimental results, we suggest to provide more comprehensive evaluations on different types of attacks:

(1) \emph{Evaluate model robustness on $\ell_p$-norm bounded adversarial attacks}. However, as shown in Table \ref{table:white-box} and \ref{table:black-box}, most adversarial
defenses are designed to counteract a single type of perturbation (\eg, small $\ell_{\infty}$-noise) and offer no guarantees for other perturbations (\eg, $\ell_1$, $\ell_2$), sometimes even increase model vulnerability \cite{Tramer2019Adversarial}. Thus, to fully evaluate adversarial robustness, we suggest to use $\ell_1$, $\ell_2$, and $\ell_{\infty}$ attacks.

(2) \emph{Evaluate model robustness on adversarial attacks as well as corruption attacks}. In addition to adversarial examples, corruption such as snow and blur also frequently occur in the real world, which also presents critical challenges for the building of strong deep learning models. According to our studies, deep learning models behave distinctly subhuman to input images with different corruption. Meanwhile, adversarially robust models may also vulnerable to corruption as shown in Fig \ref{fig:mCE}. Therefore, we suggest to take both adversarial robustness and corruption robustness into consideration, when measuring the model robustness against noises.

(3) \emph{Perform black-box or gradient-free adversarial attacks.} Black-box attacks are effective to elaborating whether obfuscated gradients \cite{athalye2018obfuscated} have been introduced to a specific defense. Moreover, black-box attacks are also shown to cover more neurons when perform test as shown in Table \ref{table:KMNCov}. However, it seems that transfer-based black-box attacks are not a good indicator for robustness evaluation, especially there exists huge-differences between source and targets models or testing on large datasets.

\subsection{Evaluate Model Robustness Considering Multiple Views}

To mitigate the problem brought by incomplete evaluations, we suggest to evaluate model robustness using more rigorous metrics, which consider multi-view robustness.

(1) \emph{Consider model behaviors with respect to more profound metrics, \eg, prediction confidence.} For example, though showing high adversarial accuracy, SAT are vulnerable by showing high confidence of adversarial classes and low confidence of true classes, which are similar to vanilla models.

(2) \emph{Evaluate model robustness in terms of model structures, \eg, boundary distance.} For example, though ranking high among other baselines on adversarial accuracy, TRADES is not strong enough in terms of Neuron Sensitivity, EBD compared to other baselines.

\subsection{Design of Attacks and Defenses}

Besides model robustness evaluation, the proposed metrics are also beneficial to the design of adversarial attacks and defenses. Most of these metrics provide deep investigations of the model behaviors or structures towards noises, which can be used for researchers to design adversarial attack or defense methods. Regarding metrics in terms of model structures, we can develop new attacks or defenses by either enhancing or impairing them, since these metrics capture the structural pattern that manifests model robustness.

\subsection{Selection of proposed metrics}

{Our proposed metrics reflect the model robustness from different perspectives. According to the specific scenario, we should select the proper metric to measure the model robustness in order to improve it. For instance, CAV, CRR, CSR, COS and CCV should be applied to compare different defensive methods, and EBD should be applied to evaluate robustness of random noises. When it comes to model structure adjustment, we should examine the neuron sensitivity and neuron uncertainty. Proper metrics can depict the model robustness more clearly in suitable scenarios, and can provide more insight that is helpful for model robustness enhancement.}

In conclusion, for comprehensive robustness evaluation, we suggest authors to: (1) evaluate robustness towards different noises (adversarial noises and common corruption); (2) use both white-box and black-box attacks for evaluating adversarial robustness; and (3) evaluate robustness from different perspectives including behaviors and structures for deeper analyses.

\section{An Open-Sourced Platform}
\label{Section:opensource}
To fully support our multi-view evaluation and facilitate further research, we provide an open-sourced platform\footnote{https://git.openi.org.cn/OpenI/AISafety} based on Pytorch. Our platform contains several highlights as follows:

(1) \emph{Multi-language environment.} To facilitate the user flexibility, we support the use of language-independent models (\eg, Java, C, Python, \etc.). To achieve the goal, we establish standardized input and output systems with a uniform format with the help of Docker.

(2) \emph{High extendibility}. Our platform also supports continuous integration of user-specific algorithms and models. In other words, users are able to introduce externally personal-designed attack, defense, evaluation methods, by simply inheriting the base classes through several public interfaces. 

(3) \emph{Multiple scenarios}. Our platform integrates multiple real-world application scenarios, \eg, auto-driving, automatic check-out, interactive robots.

Our platform enjoys several advantages as follows: (1) Attacks and defenses. Our platform contains 15 adversarial attacks, 19 corruption attacks, and 10 adversarial defenses. (2) Robustness evaluation. We have 23 different evaluation metrics. (3) Static/dynamic analysis. Our platform is the only platform that could perform static and dynamic analysis. (4) Competition. Our platform could further enable users to organize or participate in competitions on our open-source platform using the embedded interfaces. (5) Multiple scenarios. Our platform could evaluate model robustness for several real-world scenarios, \eg, automatic-checkout and auto-driving using the sandbox inside. \revision{Prevailing platforms mainly focus on attack and defense algorithms, while not able to provide comprehensive robustness evaluations. Adversarial toolboxes, \eg, Foolbox \cite{Rauber2017Foolbox} and Cleverhans \cite{papernot2016cleverhans}, only have (1), and test platforms, \eg, DeepXplore \cite{pei2017deepxplore}, have (1) and (2).}

\begin{figure}[!htb]
\vspace{-0.1in}
	\centering
	\includegraphics[width=0.6\linewidth]{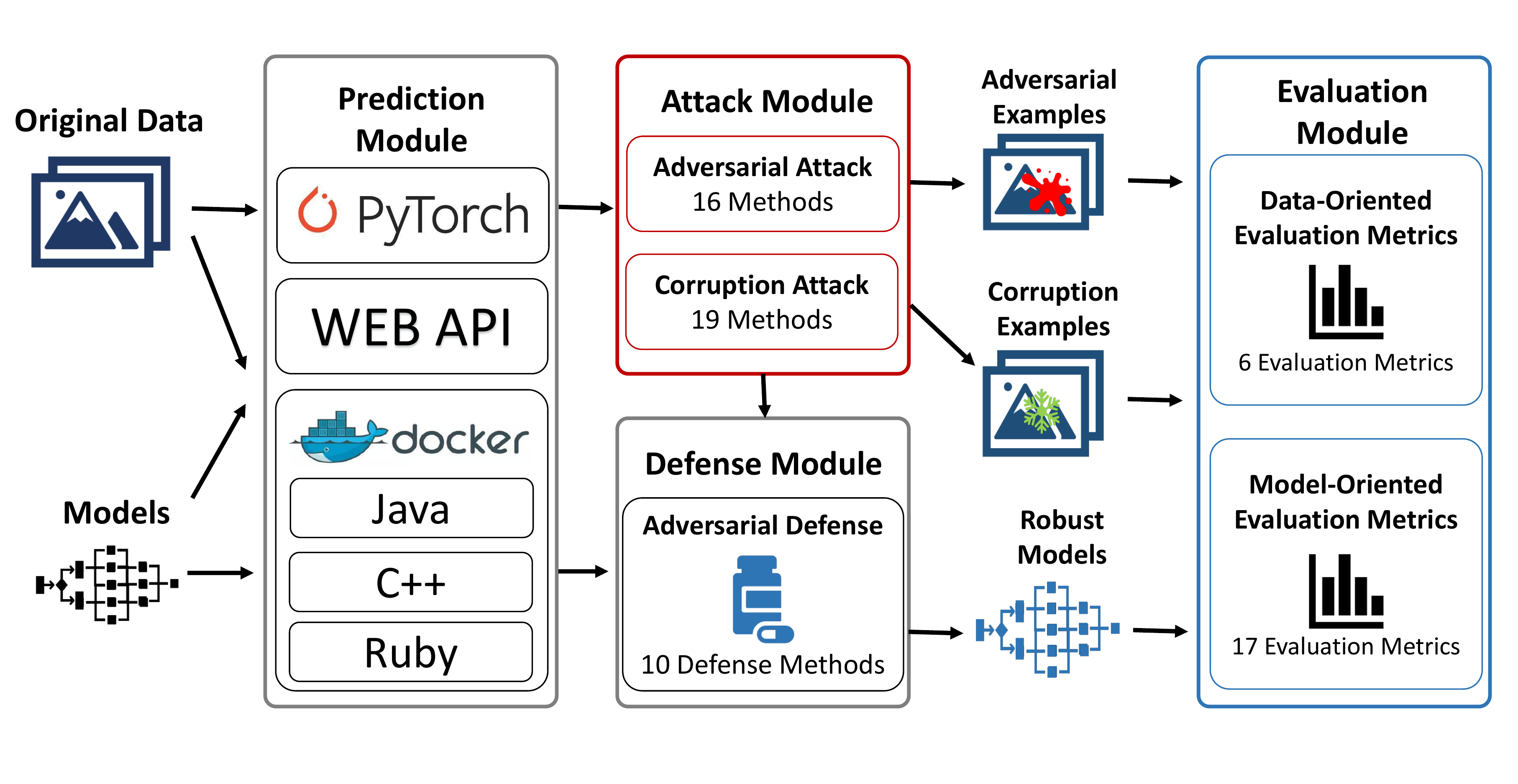}
	\caption{The framework of our open-source platform, which primarily consists of Attack module, Defense module, Evaluation module, Prediction module, and Database module.}
	\label{fig:opensource}
\vspace{-0.3in}
\end{figure}

% \begin{table}[!htb]
% \centering
% \caption{The comparison of platform and other open-sourced platforms.}
% \begin{center}
% \begin{scriptsize}
% \begin{sc}
% %\scriptsize
% \setlength{\tabcolsep}{0.3mm}{
% \begin{tabular}{cccccc}

% \toprule
%  & \tabincell{c}{Attacks\\defenses} & \tabincell{c}{Robustness\\evaluation} & \tabincell{c}{Static/dynamic\\analysis} & Competition & \tabincell{c}{Multiple\\scenarios}  \\
% \hline

% Cleverhans \cite{papernot2016cleverhans} & \checkmark & & & &  \\
% Foolbox \cite{rauber2017foolbox} & \checkmark & & & &  \\
% DeepSec \cite{Ling2019Deepsec} & \checkmark & \checkmark & & & \\
% DeepXplore \cite{pei2017deepxplore} & & \checkmark & & &  \\
% DeepTest \cite{tian2018deeptest} & \checkmark & & & & \checkmark \\
% ARES \cite{Dong2020Benchmarking} & \checkmark & \checkmark& & &\\
% Ours & \checkmark & \checkmark & \checkmark & \checkmark & \checkmark\\

% %\hline
% \bottomrule

% \end{tabular}}
% \end{sc}
% \end{scriptsize}
% \end{center}
% \label{table:opensource}
% \end{table}

\section{Conclusion}
\label{Section:conclusion}

% Current evaluations usually use simple metrics to study the performance of defenses, which are far from understanding the limitation and weaknesses of these defense methods. Thus, most proposed defenses are quickly shown to be attacked successfully, which results in the ``arm race'' phenomenon between attack and defense. To mitigate this problem, we establish a model robustness evaluation framework containing {23} comprehensive and rigorous metrics, which {consider two key perspectives of adversarial learning (\ie, data and model). To fully demonstrate the effectiveness of our framework, we conduct large-scale experiments on multiple datasets using different models and defenses with our open-source platform.}

In this work, we establish a model robustness evaluation framework containing 23 comprehensive and rigorous metrics, which consider two key perspectives of adversarial learning (\ie, data and model). {Moreover, we provide an open-sourced model robustness evaluation platform providing multiple views of model robustness with the help of the metric framework, which supports continuous integration of user-specific algorithms and language-independent models.} To fully demonstrate the effectiveness of our framework, we conduct large-scale experiments on multiple datasets using different models and defenses with our open-source platform. {The experimental results show the limit of prevailing defense methods, as they behave differently on our proposed metrics. Finally, we provide discussions and suggestions according to our experiment, hoping to shed light on model robustness.}
The objective of this work is to provide a comprehensive evaluation framework which could conduct more rigorous evaluations of model robustness. We hope our paper can facilitate fellow researchers for a better understanding of the adversarial examples as well as further improvement of model robustness. 



\bibliographystyle{main} 
\bibliography{main}

\end{document}